\documentclass[journal]{IEEEtran}
\usepackage{multirow}
\usepackage{multicol}
\usepackage{booktabs}
\usepackage{amssymb} 
\usepackage{titlesec} 

\usepackage{tabularx}
\usepackage{array}
\usepackage{booktabs}
\usepackage{xurl}
\usepackage{dblfloatfix}

\usepackage[style=ieee, backend=biber, doi=true, url=false, isbn=false,
  eprint=false,
  maxbibnames=6,    
  minbibnames=1]{biblatex}
\usepackage{placeins}

\AtEveryBibitem{
  \clearfield{abstract}
  \clearfield{keywords}
  \clearfield{note}
  \clearfield{urldate}
  \clearfield{issn}
  \clearfield{pagetotal}
  \clearlist{language}
}

\usepackage{svg}
\ifCLASSINFOpdf

  \graphicspath{{../pdf/}{../jpeg/}}
  \DeclareGraphicsExtensions{.pdf,.jpeg,.png}
\else
\fi
\usepackage{amsmath}
\usepackage{amssymb}
\usepackage[hidelinks]{hyperref}
\usepackage{orcidlink}

\begin{document}
%
\title{MotionDLO: Hybrid Event- and Frame-Based Tracking of Deformable Linear Objects}
%
%
%

\author{Annalena Hartmann\,\orcidlink{0009-0006-0283-7543},
~\IEEEmembership{Graduate Student Member,~IEEE},
Priyamvada Ajithkumar\,\orcidlink{0009-0004-1839-9352},
Patrick Br\"undl\,\orcidlink{0000-0002-3694-907X},
and~J\"org Franke\,\orcidlink{0000-0003-0700-2028},
~\IEEEmembership{Senior Member,~IEEE}
\thanks{Annalena Hartmann, Priyamvada Ajithkumar, Patrick Bründl and Jörg Franke are with the Institute for Factory Automation and Production Systems, Friedrich-Alexander-Universität Erlangen-Nürnberg, Fürther Straße 246b, 90429 Nürnberg, Germany, e-mail: annalena.hartmann@faps.fau.de}
}

%
%

\markboth{}%
{Shell \MakeLowercase{\textit{et al.}}: Bare Demo of IEEEtran.cls for IEEE Journals}
%



\maketitle

\begin{abstract}
Reliably tracking moving deformable linear objects (DLOs) while simultaneously ensuring robustness, accuracy, and temporally consistent state estimation remains a fundamental challenge in robot perception. We introduce MotionDLO, a real-time tracking framework specifically designed to overcome these limitations in temporal continuity and latency. The method exploits the high temporal resolution and sparsity of event-based cameras and combines segmentation with the Coherent Point Drift (CPD) algorithm under the principles of Motion Coherence Theory. This integration enables temporally consistent shape estimation while maintaining a low computational overhead. Existing event-based tracking methods are typically computationally efficient but exhibit reduced accuracy compared to frame-based approaches, or alternatively compromise event sparsity to achieve competitive performance. To resolve this trade-off, we propose a hybrid event- and frame-based tracking architecture that preserves the complementary strengths of both sensing modalities. The event stream ensures high-frequency motion updates, while frame-based information stabilizes spatial accuracy and object identity. We demonstrate that the proposed framework reliably associates DLO instances across video sequences, enabling robust perception for robotic manipulation tasks. Experimental results validate real-time performance at 12 ms update rates and accurate shape tracking with an point-to-curve error as measurement of accuracy of up to 0.43 mm, supporting dynamic path adaptation during manipulation. The source code and demonstration datasets are publicly available.
\end{abstract}


\section{Introduction}

The robust detection, tracking, and manipulation of DLOs constitute a prerequisite for the automation of a wide range of industrially and scientifically relevant tasks. In automotive manufacturing, for instance, the final assembly of wire harnesses into vehicle bodies remains a largely manual process that stands to benefit significantly from vision-guided robotic handling. Vision guidance extends the reach of robotic automation to tasks involving deformable, variable, or imprecisely located parts. These capabilities are essential wherever component geometry or placement cannot be tightly controlled in advance\cite{hartmann_multi-robot_2025}. Similarly, the growing scale and density of modern data center infrastructure has motivated recent work on autonomous cable routing and transceiver manipulation in cluttered rack environments \cite{sarantopoulos_robust_2025}, building on broader advances in learned robotic cable routing \cite{luo_multistage_2024}.  In the medical domain, accurate real-time tracking of suture threads is essential for enabling autonomous or semi-autonomous stitching in robot-assisted minimally invasive surgery \cite{joglekar_autonomous_2025}. Furthermore, agricultural robotics presents analogous challenges, as the structurally compliant behavior of plant branches and stems requires adaptive shape control strategies for tasks such as automated pruning and crop inspection \cite{aghajanzadeh_adaptive_2022}. Their inherent flexibility, together with properties such as low surface texture, frequent self-occlusion, entanglement, and continuously varying curvature, limits the effectiveness of conventional perception pipelines. These difficulties are further intensified in cluttered and dynamic environments commonly found in industrial applications, where DLOs overlap, intertwine, and may partially vanish from the camera’s field of view due to occlusions or poor lighting conditions. \cite{caporali_fastdlo_2022, xiang_trackdlo_2023, chi_occlusion-robust_2019, zhaole_robust_2024} 

Learning-based approaches for tracking DLOs typically require large amounts of task-specific annotated data and often exhibit limited generalization to previously unseen instances. Zero-shot segmentation foundation models have recently demonstrated the ability to reduce this annotation dependence in general robotic perception tasks \cite{zhang_zisvfm_2025}. Therefore, we propose a zero-shot approach based on Segment Anything Model 3 (SAM 3) \cite{carion_sam_2025}, combined with a CPD \cite{xiang_trackdlo_2023} tracker. This formulation eliminates the need for extensive annotated datasets for specific chosen use case while enabling robust adaptation to novel DLOs, thereby improving generalization beyond the capabilities of conventional learning-based methods. While building upon existing DLO tracking algorithms, the proposed method overcomes fundamental constraints by leveraging event-based camera data. Event-based sensing has previously demonstrated robust tracking of rapidly moving objects in general robotic contexts \cite{ni_asynchronous_2012, wang_asynchronous_2024}, and we extend this capability to the specific challenges of DLO tracking under high-speed motion. Event-based sensing has also been applied to other time-critical robotic perception tasks such as optical tactile sensing \cite{funk_evetac_2024}. Rather than replacing conventional RGB sensing, the approach complements it by extending its applicability to high-speed motion scenarios.

The proposed MotionDLO framework is built upon the non-rigid registration established by TrackDLO~\cite{xiang_trackdlo_2023}, in which CPD with a geodesic kernel propagates an ordered polyline representation across successive observations. What substantially distinguishes MotionDLO  from existing DLO trackers is the tight integration of an asynchronous event-based observation branch with a frame-based zero-shot segmentation branch within a single estimation pipeline. The event tracker accumulates polarity-separated time surfaces from the asynchronous event stream over short temporal windows and supplies the CPD tracker with binary observations within 12 ms. The frame-based branch executes SAM 3 instance segmentation within 1 s, providing low-rate but spatially accurate polylines that anchor the high-rate event-driven estimate. This arrangement preserves event sparsity for high-frequency motion updates while exploiting frame-based segmentation to stabilize spatial accuracy and maintain instance identity over long sequences, thereby resolving the latency-accuracy trade-off that has constrained prior event-based tracking approaches.

This fundamentally distinguishes MotionDLO from all SOTA approaches, which fail to adapt to large DLO deformations as demonstrated in the benchmarks. This work makes the following contributions:
 \begin{enumerate}
     \item We present \textit{MotionDLO}, a real-time tracking framework for DLOs that achieves substantially higher computational efficiency than existing methods while maintaining robustness under severe motion blur.
     \item The proposed method inherently prevents identity switches by enforcing temporally coherent correspondence estimation, enabling stable long-horizon tracking without re-identification failures.
     \item We introduce a unified tracking formulation that tightly integrates frame-based and event-based visual information within a single estimation pipeline, leveraging their complementary spatiotemporal characteristics for improved robustness and responsiveness.
     \item We provide a publicly available dataset and source code available at https://github.com/RobotDLO/MotionDLO. Extensive experimental results, including a supplementary video, demonstrate the effectiveness and reliability of the proposed approach across challenging dynamic scenarios.

 \end{enumerate}

\section{Related Work}

\begin{table*}[t]
\centering
\caption{Comparison of existing single-frame DLO segmentation and cross-frame tracking methods.}
\label{tab:dlo_methods_overview}
\scriptsize
\renewcommand{\arraystretch}{1.35}
\setlength{\tabcolsep}{4pt}

\begin{tabularx}{\textwidth}{|
>{\raggedright\arraybackslash}p{2.0cm}|
>{\raggedright\arraybackslash}p{2.8cm}|
>{\raggedright\arraybackslash}X|
>{\raggedright\arraybackslash}X|
>{\raggedright\arraybackslash}X|}
\hline
\textbf{Task} & \textbf{Category} & \textbf{Methods} & \textbf{Advantages} & \textbf{Limitations} \\
\hline

\multirow{3}{=}{Single-frame segmentation}
& Generic instance segmentation
& CNN-based: YOLACT~\cite{bolya_yolact_2019}, YOLACT++~\cite{bolya_yolact++_2022}, SOLOv2~\cite{wang_solov2_2020}, CondInst~\cite{tian_condinst_2020}; Two-stage: Mask R-CNN~\cite{he_mask_rcnn_2017}, Mask R-CNN + PointRend~\cite{dinkel_wire_2022}; Transformer: Mask2Former~\cite{cheng_mask2former_2022}, QueryInst~\cite{fang_instances_2021}
& General-purpose masks via decoupled heads, boundary refinement, or attention
& Poor fidelity on pixel-wide DLOs \newline Transformer variants are often too computationally demanding for closed-loop robotic systems with strict latency requirements. \\
\cline{2-5}
& DLO-specific graph / skeleton pipelines
& Ariadne~\cite{degregorio_ariadne_2018}, Ariadne+~\cite{caporali_ariadne_plus_2022}, FASTDLO~\cite{caporali_fastdlo_2022}, RT-DLO~\cite{caporali_rtdlo_2023}, mBEST~\cite{choi_mbest_2023}
& DLO-tailored segmentation, crossing handling, centerline extraction
& Per-frame only, no temporal propagation or cross-frame identity \\
\cline{2-5}
& Foundation-model / vision-language segmentation
& SAM family: SAM~\cite{kirillov_sam_2023}, SAM~2~\cite{ravi_sam2_2024}, SAM~3~\cite{carion_sam_2025}; Vision-language: CLIPSeg~\cite{luddecke_clipseg_2022}, ISCUTE~\cite{kozlovsky_iscute_2024}, DLO Perceiver~\cite{caporali_dlo_2024}
& Zero-shot or text-guided DLO selection; useful for initialization and drift correction
& Masks only, no centerline state or temporal identity \newline High GPU cost for real-time inference \\
\hline

\multirow{2}{=}{Cross-frame tracking}
& Registration-based
& TrackDLO~\cite{xiang_trackdlo_2023}, MultiDLO~\cite{xiang_multidlo_2023}
& Temporally coherent state via point-set registration and Motion Coherence Theory
& Degrade under fast motion \newline RGB-D bound and capped at camera frame rate \\
\cline{2-5}
& Learning-based
& UniStateDLO~\cite{lv_unistatedlo_2025}
& Cross-frame state estimation via learned generative diffusion
& Complex multi-stage RGB-D / point-cloud pipeline \\
\hline

\end{tabularx}
\end{table*}
\vspace{-4pt}
\subsection{Real-Time Instance Segmentation}

Visual perception is central to estimating object boundaries, geometry, and motion in robotic manipulation. For deformable linear objects (DLOs), high shape variability, weak visual distinctiveness, and perceptual ambiguity make instance-level recognition particularly difficult. Since closed-loop manipulation depends on continuous state feedback, perception latency directly limits the achievable control bandwidth, making real-time segmentation a fundamental requirement.

General-purpose instance segmentation methods offer useful baselines but remain poorly matched to thin DLOs. Single-stage architectures such as YOLACT~\cite{bolya_yolact_2019}, YOLACT++~\cite{bolya_yolact++_2022}, 
SOLOv2~\cite{wang_solov2_2020}, and CondInst~\cite{tian_condinst_2020} achieve near real-time throughput by decoupling classification from mask generation, yet restricted kernels, limited channel capacity, or fixed 
global prototypes reduce their fidelity for structures spanning only a few pixels. Mask~R-CNN~\cite{he_mask_rcnn_2017} similarly predicts fixed-resolution instance masks, poorly suited to thin elongated objects; Dinkel~et~al.~\cite{dinkel_wire_2022} augmented it with a PointRend head for RGB-D wire segmentation, but the underlying resolution ceiling persisted. Transformer-based methods provide stronger contextual modeling at greater computational cost: Mask2Former~\cite{cheng_mask2former_2022} 
achieves state-of-the-art accuracy through masked cross-attention and multi-scale feature processing, reaching only 4--10~fps depending on backbone, while QueryInst~\cite{fang_instances_2021} improves mask quality via parallel supervision on dynamic mask heads conditioned on instance queries, achieving 10--13~fps under practical configurations. Neither meets the latency requirements of closed-loop robotic control. These 
results indicate that richer representations improve thin-object segmentation on general benchmarks, but real-time deployment remains the primary constraint for robotic manipulation.

In the DLO-specific literature, a sequence of progressively refined processing pipelines has improved both efficiency and accuracy. Ariadne~\cite{degregorio_ariadne_2018} introduced a superpixel graph-based instance tracing strategy, which was later extended in Ariadne+~\cite{caporali_ariadne_plus_2022} through the integration of deep semantic segmentation priors; both methods remain limited to per-frame processing without temporal state propagation. FASTDLO~\cite{caporali_fastdlo_2022} couples morphological skeletonization with a learned pairwise similarity network for intersection disambiguation, achieving 22 to 23 fps on $640\times 360$ images alongside 97.4\% intersection-layout accuracy on a curated test set; its reliance on skeleton-based topology, however, makes it vulnerable to mask degradation and produces unreliable results on configurations with near-parallel crossings or self-loops. RT-DLO~\cite{caporali_rtdlo_2023} reformulated DLO segmentation as a graph-based inference over detected centerline nodes with cosine-similarity orientation scoring, surpassing 30 fps while improving per-frame intersection over union (IoU) by 2-3 percentage points relative to FASTDLO and increasing robustness to eroded segmentation masks; however, its sparse centerline sampling can be insufficient for highly variable curvature, where a coarse vertex density cannot faithfully represent the underlying geometry. mBEST~\cite{choi_mbest_2023} further improves runtime efficiency 
through minimal bending-energy skeleton traversal, achieving competitive throughput on geometrically complex configurations. However, its skeleton-based formulation remains sensitive to incomplete or ambiguous topologies, and does not explicitly resolve cases involving occlusions, multiple DLOs at a single intersection, or densely knotted configurations. A critical limitation shared by all these methods is their per-frame processing paradigm: no temporal state is propagated across frames, and RT-DLO explicitly acknowledges this lack of cross-frame instance tracking and identifies its integration as an important direction for future work. UniStateDLO~\cite{lv_unistatedlo_2025} addressed cross-frame state estimation via a learned generative diffusion framework, yet its multi-stage RGB-D point-cloud pipeline introduces substantial computational overhead and performance degradation under fast DLO motion. 

Foundation models have introduced zero-shot segmentation capabilities relevant to DLO perception. SAM~\cite{kirillov_sam_2023} generalizes across object categories but often produces coarse boundaries on thin structures. SAM~2~\cite{ravi_sam2_2024} extends SAM to video using streaming memory and supports higher throughput, although fidelity degrades for fine structures under fast motion. SAM~3~\cite{carion_sam_2025} introduces concept-prompted segmentation via a decoupled recognition-localization architecture, enabling text-conditioned masks at approximately 30 ms per frame. These 30 ms are reached on NVIDIA H200, which are not suitable for robotic use-cases, limiting direct deployment on resource-constrained robotic platforms. However, its improved boundary precision and zero-shot generalizability make it a promising backbone for DLO perception, especially given the limited transfer of task-specific segmentation models to unseen object categories without adaptation~\cite{lu_self-supervised_2023}. 

Vision-language approaches further bridge semantic understanding with spatial segmentation. CLIPSeg~\cite{luddecke_clipseg_2022} enables text-conditioned mask prediction, but its CLIP ViT-B/16 backbone downsamples inputs and reduces thin structure detail, and producing a single foreground/background mask rather than separating co-present instances. ISCUTE~\cite{kozlovsky_iscute_2024} addressed the latter by coupling CLIPSeg with SAM through a learned adapter, attaining state-of-the-art DLO mean IoU on its own benchmark; however, the pipeline relies on two frozen foundation backbones ($\approx$ 466 million parameters) and inherits their failure modes, with the authors reported that CLIPSeg occasionally fails to localize low-contrast cables and that the adapter's mask classifier discards valid submasks in cluttered scenes. DLO~Perceiver~\cite{caporali_dlo_2024} introduced a DLO-specific language-guided segmentation framework, using a compact perceiver-inspired attention module to fuse image and prompt embeddings and a contrastive branch to estimate image-prompt consistency. Nevertheless, each inference targets the DLO specified by a structured object-color-position prompt, and the performance depends on accurate prompt specification, thereby requiring reliable target-specific attribute information during deployment. Across all three methods, the output remains a single-frame mask without the propagation of centerline state or instance identity over time, leaving the temporal continuity required for closed-loop manipulation unaddressed.

Despite these advances, a gap remains between accurate per-frame segmentation and temporally coherent DLO perception for manipulation. Thin DLOs are strongly affected by downsampling and fixed-resolution mask prediction, while weak texture, clutter, and crossings make instance boundaries ambiguous. TrackDLO addresses temporal continuity through point-set registration under Motion Coherence Theory (MCT), but its dependence on RGB-D sensing limits applicability when DLO diameters fall below reliable depth resolution, and its small-displacement assumption can lead to correspondence errors under fast motion. UniStateDLO provides learning-based cross-frame tracking, but requires a computationally demanding RGB-D pipeline and extensive training. SAM-based DLO segmentation has also been explored as an observation front-end~\cite{zhaole_robust_2024}, but segmentation alone does not resolve temporal continuity. Existing shape-estimation methods therefore remain largely focused on detecting or segmenting DLOs in individual frames~\cite{caporali_robotic_2025}, leaving robust temporally consistent tracking under fast motion unresolved.The reviewed segmentation and tracking approaches, together with their main sensing modalities, advantages, and limitations, are summarized in Table \ref{tab:dlo_methods_overview}.

\subsection{Event-based Tracking}

Event cameras provide a complementary sensing modality for addressing this limitation. As it asynchronously report per-pixel log-intensity changes with microsecond latency, $>$120 dB dynamic range, and sparse, edge-centric output, making them inherently suited to low-latency, motion-blur-free tracking under fast dynamics and challenging illumination conditions~\cite{gallego_event-based_2022, gehrig_low-latency_2024}.
Event-based tracking has evolved from low-level feature association toward learning-based, object-level frameworks. Early approaches processed the event stream directly through feature-based methods such as asynchronous corner detection and tracking~\cite{alzugaray_asynchronous_2018}, or motion-compensation and clustering strategies for independent moving-object detection~\cite{mitrokhin_event-based_2018}. In parallel, dense motion estimation via learned optical flow from events, exemplified by EV-FlowNet~\cite{zhu_ev-flownet_2018} and E-RAFT~\cite{gehrig_e-raft_2021}, enabled temporally consistent warping and alignment, forming the basis for higher-level tracking pipelines.
More recently, learning-based methods have expanded the representational palette by ingesting voxel grids, time surfaces, or accumulated event images into CNN, transformer, or spiking-network architectures. Hybrid frame–event feature trackers~\cite{messikommer_data-driven_2025} combine the spatial richness of frames with the temporal density of events, while fully spiking fusion trackers such as SpikeFET~\cite{yang_fully_2025} target energy-efficient inference on neuromorphic hardware. Dedicated benchmarks including VisEvent~\cite{wang_visevent_2023} and EventVOT~\cite{wang_event_2024} further attest to the growing maturity of the field.
Despite this progress as described in Table~\ref{tab:event_methods_overview}, a persistent trade-off remains: methods that operate directly on the sparse event stream achieve low latency but often sacrifice spatial precision, whereas approaches that convert events into dense frame-like representations recover accuracy at the cost of the very sparsity that makes event cameras attractive. Furthermore, existing event-based trackers have predominantly targeted rigid-object or generic bounding-box tracking; applications to deformable structures remain scarce. A notable exception is the work of Panetsos et~al.~\cite{panetsos_aerial_2024}, who leverage event-camera data and B\'{e}zier curve fitting to estimate the state of a cable-suspended load during aerial transportation. However, their method addresses a single, quasi-rigid cable under constrained geometric assumptions and does not generalize to multi-instance DLO tracking in cluttered manipulation scenes. More broadly, open-vocabulary event-based perception has recently advanced with SEAL ~\cite{lee_seal_2026}, which proposes an efficient multimodal segmentation framework with strong benchmark results at 22ms per frame; nonetheless, its training and evaluation are confined to outdoor driving scenes containing no thin or deformable structures, making it insufficient for DLO tracking in manipulation scenarios. Dietsche~et~al.~\cite{dietsche_powerline_2021} demonstrated the viability of event-based tracking for thin linear structures in outdoor powerline inspection scenarios; however, their approach targets a single taut cable 
in an uncluttered outdoor environment and does not generalize to 
multi-instance DLO tracking in cluttered manipulation scenes. More broadly, no existing method integrates event-camera data with RGB-based instance segmentation within a unified framework for multi-DLO manipulation.
These limitations motivate the hybrid event- and frame-based architecture proposed in this work, which preserves event sparsity for high-frequency motion updates while relying on frame-based segmentation to stabilize spatial accuracy and maintain instance identity across time.

\begin{table*}[t]
\centering
\caption{Overview of event-based segmentation and tracking methods relevant to MotionDLO.}
\label{tab:event_methods_overview}
\scriptsize
\renewcommand{\arraystretch}{1.35}
\setlength{\tabcolsep}{4pt}

\begin{tabularx}{\textwidth}{|
>{\raggedright\arraybackslash}p{2.0cm}|
>{\raggedright\arraybackslash}p{2.8cm}|
>{\raggedright\arraybackslash}X|
>{\raggedright\arraybackslash}X|
>{\raggedright\arraybackslash}X|}
\hline
\textbf{Task} & \textbf{Category} & \textbf{Methods} & \textbf{Advantages} & \textbf{Limitations} \\
\hline

\multirow{3}{=}{Motion \& feature estimation}
& Event-camera sensing
& Gallego et al.~\cite{gallego_event-based_2022}, Gehrig et al.~\cite{gehrig_low-latency_2024}
& Microsecond latency, HDR, blur-robust edge output
& Raw events lack DLO masks or ordered state \\
\cline{2-5}
& Event-based feature / object trackers
& Async corners: Alzugaray et al.~\cite{alzugaray_asynchronous_2018}; Moving-object detection: Mitrokhin et al.~\cite{mitrokhin_event-based_2018}; Hybrid frame-event: Messikommer et al.~\cite{messikommer_data-driven_2025}; Spiking fusion: Yang et al.~\cite{yang_fully_2025}
& Fast feature- or object-level tracking; hybrid and spiking variants add blur robustness and neuromorphic efficiency
& Sparse features or generic moving objects, not DLO geometry or identity \\
\cline{2-5}
& Event optical flow
& EV-FlowNet~\cite{zhu_ev-flownet_2018}, E-RAFT~\cite{gehrig_e-raft_2021}, time-surface matching~\cite{nagataOpticalFlowEstimation2021}, triplet matching~\cite{shibaFastEventBasedOptical2022}
& High-rate motion fields for warping under fast motion
& No DLO instance identity or ordered centerline \\
\hline

\multirow{2}{=}{Object / structure segmentation}
& Event-based thin-structure tracking
& Powerlines: Dietsche et al.~\cite{dietsche_powerline_2021}; Cable-suspended loads: Panetsos et al.~\cite{panetsos_aerial_2024}
& Event-based tracking of thin, cable-like structures
& Constrained setups, not cluttered multi-DLO manipulation \\
\cline{2-5}
& Open-vocabulary event perception
& SEAL~\cite{lee_seal_2026}
& Multimodal open-vocabulary event segmentation
& Outdoor driving benchmarks, not DLO manipulation \\
\hline

\end{tabularx}
\end{table*}

\section{The MotionDLO Algorithm}

\subsection{Problem Formulation}

\subsubsection{Operation Conditions and Constraints}

In an industrial context, DLOs can have diameters as small as 1 mm. A region of interest of approximately 200 mm × 200 mm is taken as a reference, which can be covered by a standard RGB or event-based camera with a resolution of 1280 × 720 px at a working distance of approximately 200 mm, which is a typical distance for a flange-mounted camera on an industrial robot. To establish a baseline for the required temporal resolution, the DLO is assumed to start from a static position and undergo free-fall acceleration at 9.81 m/s². Under these conditions, the DLO traverses the 200 mm field of view in approximately 200 ms. At a conventional frame rate of 50 fps, this corresponds to 10 captured frames. However, because the DLO continuously accelerates, the inter-frame displacement grows throughout the transit. At the bottom of the field of view the DLO reaches a velocity of approximately 2000 mm/s, corresponding to a displacement of roughly 40 mm between consecutive frames exceeding the DLO diameter by a factor of 40.

\begin{figure*}
    \centering
    \includegraphics[width=1\textwidth]{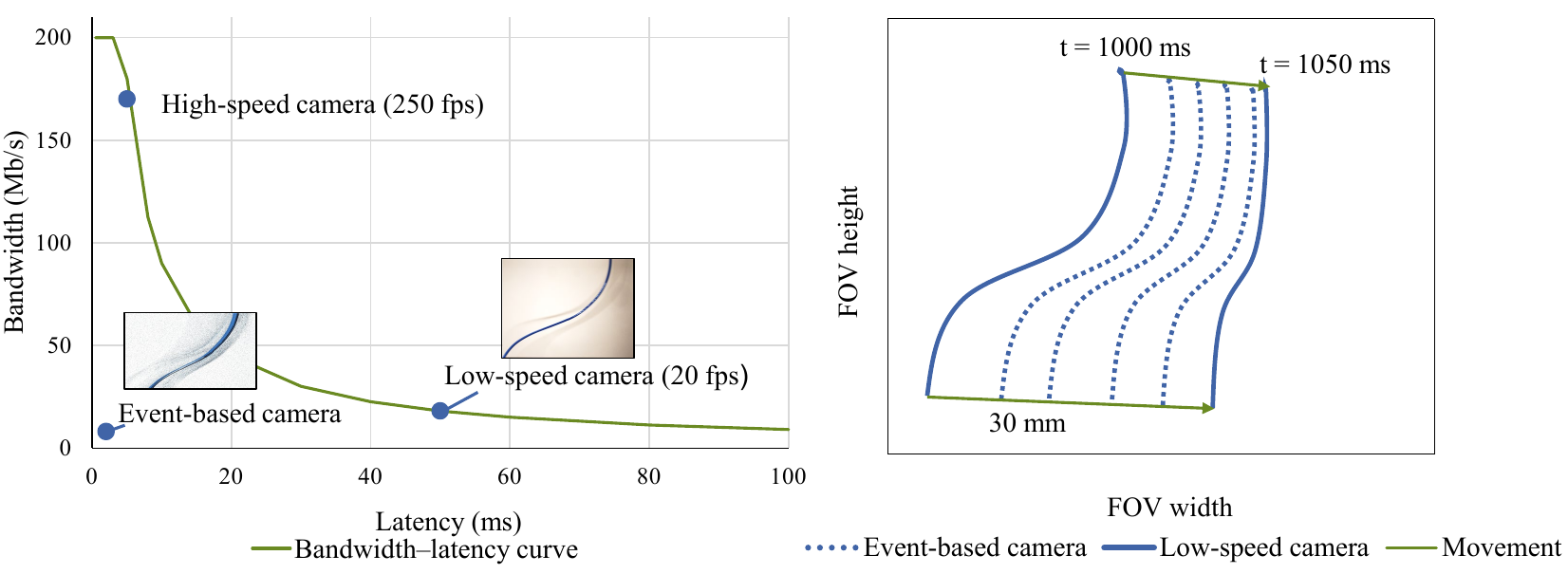}
    \caption{Bandwidth–latency trade-off for DLO observation. \textit{(Left)} Along the Pareto front (green), a 250 fps high-speed camera attains low latency at the cost of Mb/s, while a 20 fps frame camera remains bandwidth-efficient but incurs 50 ms latency and pronounced motion blur. The event-based camera combines sub-millisecond latency with an order-of-magnitude lower bandwidth than the high-speed camera while yielding a sparse, blur-free signal. \textit{(Right)} Within a single 50 ms inter-frame interval of the low-speed camera, the DLO traverses 30 mm across the field of view, a displacement that defeats CPD-based frame trackers. The event stream resolves this motion as a continuous sequence of intermediate shapes.}
    \label{fig:latency-bandwith}
\end{figure*}

 A second scenario would be the movement of the DLO with an industrial robot, which reaches speeds of up to 5000 mm/s and 2000 mm/s for cobots. This scenario is depicted in figure~\ref{fig:latency-bandwith}. The moving DLO traverses inconsistently through the image by 30 mm within 50 ms. Even moderate displacements of a few DLO diameters are sufficient to cause failure in point-set registration methods such as CPD and lead to identity switches in multi-object trackers like RT-DLO. The maximum of TrackDLO as the most advanced CPD method is for instance 150 mm/s due to small inter-frame movement of maximum 5 mm and 30 frames per second. A higher fps camera would not solve the problem as computing time is crucial and a trade-off between bandwidth and latency is present as depicted in figure~\ref{fig:latency-bandwith} \cite{gehrig_low-latency_2024}. Consequently, existing frame-based approaches lack the temporal resolution required for reliable DLO tracking, particularly in dense environments where multiple DLOs are present.

\subsubsection{Problem Statement} \label{Problem Statement}
A scene contains $K$ deformable linear objects. Although each DLO physically resides in three-dimensional space, the present work restricts the estimation to two-dimensional projections in the event sensor pixel frame; an extension to three dimensions is discussed in Section~\ref{Future Work}. Each DLO instance $i \in \{1,\dots,K\}$ is characterized by its length $l^{(i)}$ and approximately constant cross-sectional width $w^{(i)}$. Its state at time $t$ is represented as an ordered chain of $M_i$ control nodes,
\begin{equation}
    Y_t^{(i)} = \bigl[\mathbf{y}_{1,t}^{(i)},\, \mathbf{y}_{2,t}^{(i)},\, 
    \dots,\, \mathbf{y}_{M_i,t}^{(i)}\bigr]^{\top} \in \mathbb{R}^{M_i \times 2},
    \label{eq:state}
\end{equation}
where the node order follows the physical continuity of the DLO along its 
arc length. Each DLO is assumed to be approximately inextensible, so that the total arc length $L_t^{(i)} = \sum_{m=2}^{M_i} \lVert \mathbf{y}_{m,t}^{(i)} - 
\mathbf{y}_{m-1,t}^{(i)} \rVert_2$ satisfies $L_t^{(i)} \approx l^{(i)}$ for all~$t$.

At each time $t$, every instance~$i$ occupies a motion state $z^{(i)}_t \in \{0,1\}$, where $z^{(i)}_t = 0$ denotes a static DLO and $z^{(i)}_t = 1$ denotes an actively deforming DLO. The motion state is a property of the underlying scene and is independent of the perception modality used to estimate it. The DLO perception problem consists of estimating, for every instance $i$ and at every time $t$, an estimate $\hat{Y}_t^{(i)}$ of the ground-truth state $Y_t^{*(i)}$ such that
\begin{equation}
    \bigl\lVert \hat{Y}_t^{(i)} - Y_t^{*(i)} \bigr\rVert
    \label{eq:estimation_ground_truth}
\end{equation}
is minimized, given partial and noisy observations arising from occlusions, imperfect segmentation, and motion blur, and without recourse to explicit physical priors. Identity preservation requires that, for every instance~$i$ tracked at time~$t-1$ and still present at~$t$, the estimate $\hat{Y}_t^{(i)}$ corresponds to the same physical DLO as $\hat{Y}_{t-1}^{(i)}$. An identity switch denotes any event in which a tracked instance is reassigned to a different physical DLO or fragmented across instance labels.

The hybrid tracking problem addressed in this work is to produce $\{\hat{Y}_t^{(i)}\}_{i=1}^{K}$ at event-camera update rates while preserving instance identity over arbitrarily long sequences, given asynchronous event observations and lower-rate frame observations of the same scene.

\subsection{Overall Pipeline}

\begin{figure*}
    \centering
    \includegraphics[width=\textwidth]{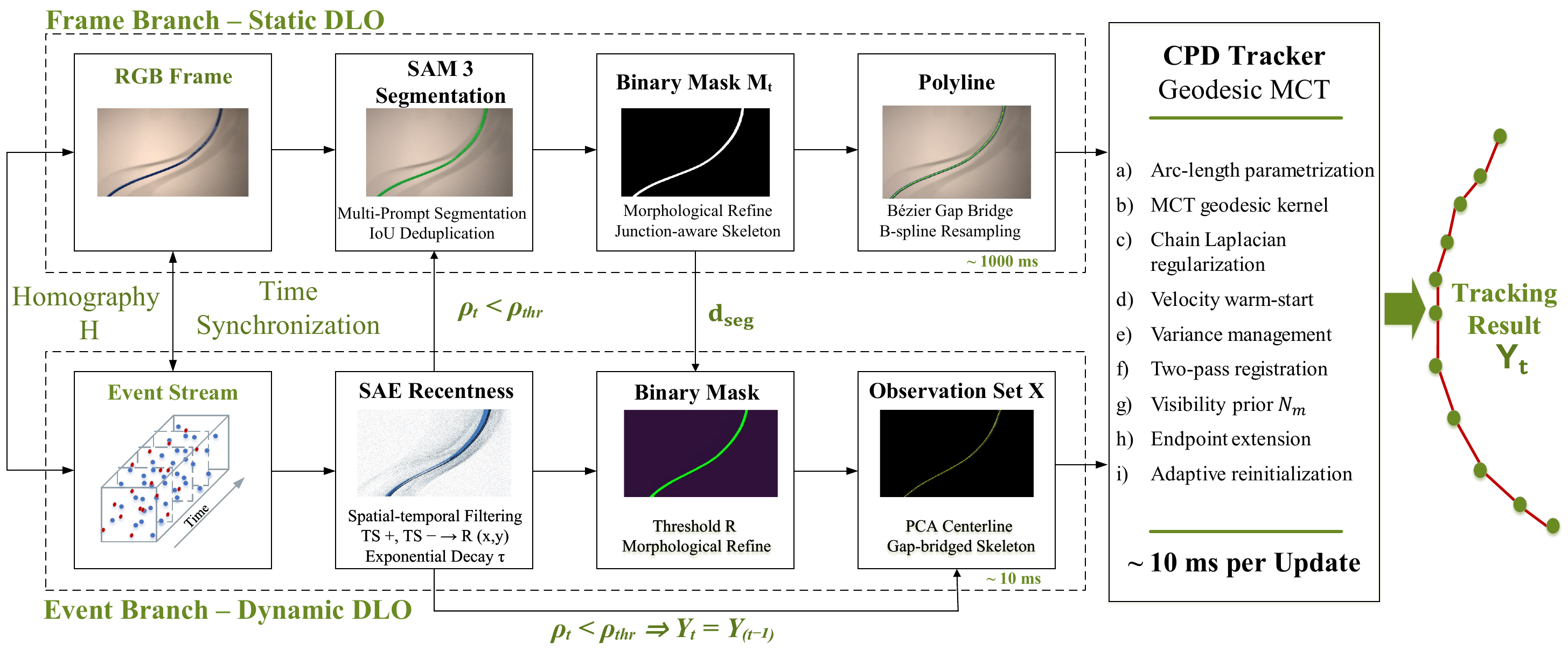}
    
    \caption{The tracking result $Y_t$ is computed at 12\,ms intervals by the CPD tracker operating on a geodesic MCT representation. At each update, the tracker receives polyline input from one of two branches: the frame-based branch~(A) or the event-based branch~(B). Branch selection depends on the DLO motion state, which is determined by the threshold parameter $\rho_{\mathrm{thr}}$ defined as the count of DLO-attributed events within a fixed temporal window. A motion state of $v_{\mathrm{DLO}} = 0$ triggers branch~(A), while $v_{\mathrm{DLO}} \neq 0$ triggers branch~(B). Temporal alignment between the two branches is ensured via hardware trigger timestamps; spatial alignment is established through the homography~$H$.}
    \label{fig:Overview}
\end{figure*}

The core idea of the proposed approach relies on detecting the DLOs and segmenting them with regular frame-based RGB cameras while generating fast intermediate updates with event-based cameras, in order to overcome the temporal resolution and motion blur limitations inherent to purely frame-based tracking methods. The corresponding implementation depicts this hybrid architecture as a two-branch pipeline as presented in figure~\ref{fig:Overview}. In the frame-based branch, running in a static scenario while no DLO motion is detected, an RGB frame is passed to a multi-prompt SAM 3 zero-shot segmentation module, queried with several DLO-related text prompts (e.g.\ \textit{cable}, \textit{DLO}, \textit{hose}) to maximize recall, which produces a merged binary mask from which a high-accuracy polyline and DLO parameters for tuning the event-based branch are extracted. In the event-based branch, operating at 5 ms update intervals, raw events from an event camera are first processed through a spatiotemporal filter to suppress noise then accumulated into polarity-separated time surfaces from which a recentness map is computed using a Surface of Active Events (SAE) representation with an adaptive temporal decay. A binary DLO mask is derived by thresholding this recentness map and refined via morphological closing and opening operations; the resulting mask is skeletonized to extract a centerline, which serves as the observation point set for a CPD-based geodesic curve tracker. This tracker maintains a parametric polyline of equidistantly resampled nodes whose deformation is governed by a second-order smoothing kernel evaluated over geodesic distances along the curve, thereby enforcing the motion coherence constraint during iterative Expectation–Maximization registration. The two branches are fused with hardware triggered time-stamping and calibrated polylines: the frame worker posts confidence- and timestamp-weighted correction signals that the event loop integrates into the tracker state, thereby combining the high temporal responsiveness of the event path with the spatial accuracy and identity stability of the frame path.

The proposed framework extends motion-coherence-based DLO tracking to operate purely on RGB imagery. In contrast to the TrackDLO algorithm, which relies on RGB-D sensing and 3D point cloud registration, the proposed method eliminates the dependency on depth information and instead leverages high-quality segmentation masks extracted from RGB frames. This enables reliable tracking of thin DLOs in scenarios where depth sensing is unreliable or noisy.

\subsection{Frame Branch: Static DLO}

\paragraph{SAM 3 Segmentation}

To obtain reliable observations of the DLO in each RGB frame, the SAM~3 foundation segmentation model with text-prompt 
conditioning is employed~\cite{carion_sam_2025}. Instead of 
relying on a single semantic query, multiple DLO-related prompts (e.g., \textit{cable}, \textit{wire}, \textit{hose}, 
\textit{tubing}) are used to improve recall under varying appearance conditions. Since SAM~3 inference latency precludes 
real-time deployment on robotics hardware, segmentation is restricted to the static DLO state, where latency is not critical.

For each static RGB frame, SAM 3 is evaluated independently for each prompt, producing a set of candidate binary masks. Each mask is thresholded at a fixed sigmoid confidence value and discarded if its resulting foreground region falls below a minimum threshold. To prevent duplicate detections across prompts, the retained masks are sorted by area in descending order and de-duplicated by Intersection-over-Union,
\begin{equation}
\mathrm{IoU}(m_i, m_j) = \frac{|m_i \cap m_j|}{|m_i \cup m_j|},
\end{equation}
where $m_i$ and $m_j$ are any two candidate masks.

A mask is suppressed if its IoU with any already-accepted mask exceeds a predefined overlap threshold, ensuring that the larger of two overlapping detections is always retained. All accepted masks are then merged into a single binary segmentation mask $M_t$ via pixel-wise union,

\begin{equation}
M_t = \bigcup_k m_k,
\end{equation}
where each $m_k$ is a binarized candidate mask retained after de-duplication.

\paragraph{Binary Mask Generation}
The merged mask is subsequently refined through a sequence of morphological operations before polyline fitting. First, binary closing is applied using a disk-shaped structuring element. Since binary closing consists of dilation followed by erosion, it connects small breaks in the foreground mask while approximately preserving the object boundary. This step reduces local discontinuities caused by imperfect segmentation, low contrast, or specular reflections. Small connected components that typically correspond to isolated segmentation artifacts rather than valid DLO regions are then removed and small holes inside the DLO region are filled to obtain a solid foreground mask. A final light closing operation further consolidates the cleaned mask. Remaining blobs below a minimum final area threshold are discarded, yielding the refined binary mask $M_t$. The refined binary mask $M_t$ represents the frame-based DLO observation obtained during static phases, i.e., when no relevant DLO motion is detected. In MotionDLO, this observation is used to derive a high-accuracy geometric reference for the hybrid tracker. 

\label{polyline fitting}
 
\paragraph{Polyline fitting}

The polyline fitting stage extracts from $M_t$ an ordered centerline model of the DLO, yielding a structured curve representation that is compatible with the subsequent CPD-based tracking and fusion stages. The resulting polyline is expressed as a fixed-length sequence of equidistant nodes,
\begin{equation}
Y_t =
[
\mathbf{y}_{1,t},
\mathbf{y}_{2,t},
\ldots,
\mathbf{y}_{M,t}
]^\top ,
\end{equation}
where the node order follows the physical continuity of the DLO along its arc length. This frame-based polyline provides a spatially aligned reference state for tracker initialization, accumulated-drift correction, and synchronization with the high-rate event-based estimates.

Since morphological thinning preserves the topology of the input mask, any unbridged gap at this stage produces a fragmented skeleton that must subsequently be reconnected. A second closing pass is therefore applied prior to thinning, with a disk radius selected adaptively. The minimum pairwise distance between connected components is examined, and when this distance falls below twice the default radius, indicating that components lie within the closing reach of one another, a reduced radius is applied to preserve their separation; otherwise, the full radius is used to robustly close intra-DLO discontinuities.

Following closing, components below a small empirically chosen area are discarded, as such isolated regions almost always correspond to segmentation noise rather than genuine DLO fragments. Each surviving component is then reduced to a one-pixel-wide medial axis by iterative morphological thinning, in which boundary pixels are peeled away while preserving the connectivity of the foreground region. The resulting skeleton is represented as an undirected graph in which every foreground pixel is a node connected to its 8-connected neighbors, so that adjacent skeleton pixels (including diagonals) form edges. The number of incident edges, or degree, at each node distinguishes endpoints (degree one), interior pixels (degree two), and junctions or crossings , providing the structural basis for the junction-aware traversal described next.

Skeleton traversal is made junction-aware to prevent distinct DLOs from being merged at crossing points. At each junction, incident branches are paired by maximizing the collinearity of their tangent directions, estimated from the first pixels along each branch,

\begin{equation}
\underset{\text{pairing}}{\arg\max} \sum_{(a,\,b)} \bigl|\hat{\mathbf{t}}_a \cdot \hat{\mathbf{t}}_b\bigr|,
\end{equation}

where $\hat{\mathbf{t}}_a$ and $\hat{\mathbf{t}}_b$ are the unit tangents of branches $a$ and $b$, estimated from the first pixels along each branch, over all valid one-to-one pairings of incident branches.

The resolved pairings constrain graph traversal so that at any junction the walker continues along its pre-assigned collinear partner. Ordered pixel sequences are extracted by walking from each tip through the constrained graph until another tip or a visited edge is reached. Since SAM 3 operates as a class-agnostic segmentation model, it produces category-level masks but cannot distinguish between individual DLO instances in multi-DLO scenes. If overlapping of DLOs occurs, no distinction is made. This overlapping can currently only be resolved by the event based branch if one of the DLOs moves and then an reinitialization on the event stream is conducted. 

Partial occlusion or low local contrast may produce multiple disjoint segments belonging to a single DLO. All pairs of segment endpoints are therefore evaluated as connection candidates, subject to a maximum gap threshold $d_{\max}$ and a minimum tangent alignment criterion. Each candidate connecting endpoints $\mathbf{e}_i$, $\mathbf{e}_j$ with 
outward tangents $\hat{\mathbf{d}}_i$, $\hat{\mathbf{d}}_j$ directed toward 
the gap is scored by,
\begin{equation}
s_{ij} = -(\hat{\mathbf{d}}_i \cdot \hat{\mathbf{d}}_j) \cdot 
\left(1 - \sqrt{\frac{\|\mathbf{e}_i - \mathbf{e}_j\|}{d_{\max}}}\right)
\end{equation}

jointly rewarding tangent alignment and penalizing larger gaps. Segments are merged greedily in descending score order using a union-find structure, with each endpoint permitted at most one connection. Accepted gaps are bridged by cubic B\'{e}zier curves with endpoints 
$\mathbf{e}_i$, $\mathbf{e}_j$ and interior control points
\begin{equation}
\mathbf{c}_1 = \mathbf{e}_i + \alpha\,\hat{\mathbf{d}}_i, \qquad 
\mathbf{c}_2 = \mathbf{e}_j - \alpha\,\hat{\mathbf{d}}_j,
\end{equation}
where $\hat{\mathbf{d}}_i$ is the outward tangent at $\mathbf{e}_i$ directed 
toward the gap, $\hat{\mathbf{d}}_j$ is the inward tangent at $\mathbf{e}_j$ 
directed into its segment, and $\alpha = 0.4\|\mathbf{e}_j - \mathbf{e}_i\|$.

The concatenated point sequence is smoothed by a Savitzky-Golay filter applied independently to each coordinate axis, suppressing the pixel-grid staircase artifacts of the discrete skeleton without displacing the curve from the DLO centerline. The $K$ smoothed skeleton points $\left\{\mathbf{p}_k\right\}_{k=1}^{K}$ are 
then resampled to $M$ equidistant nodes via chord-length parameterization, where each skeleton point $\mathbf{p}_i$ is assigned a normalized arc position

\begin{equation}
    u_i = \frac{\displaystyle\sum_{k=1}^{i-1}\|\mathbf{p}_{k+1} - \mathbf{p}_k\|}{L}, 
    \qquad i = 1, \ldots, K,
\end{equation}
and $L = \sum_{k=1}^{K-1}\|\mathbf{p}_{k+1} - \mathbf{p}_k\|$ is the total arc length of the smoothed sequence. The normalized parameters $\{u_i\}$ are then used as parameter values in a cubic B-spline fit, yielding the final polyline. $Y_t = \{\mathbf{y}_1, \ldots, \mathbf{y}_M\}$.  

Based on the segmented mask, the DLO diameter $d_{seg}$ is further estimated as twice the 75th percentile of the distance-transform values inside $M_t$ and warped into event-sensor space using the same arc-length scaling applied to the polyline, yielding a robust, outlier-resistant width estimate

When a homography ${H}$ between the RGB and event camera frames is available, the polyline node coordinates are warped into event sensor space via the projective map $\hat{\mathbf{y}}_m = \mathcal{P}(H,\, \mathbf{y}_m)$; this point-level warping avoids the rasterization artifacts that arise when warping the binary mask directly.

\subsection{Event Branch: Dynamic DLO}
 
In the event-based representation, each DLO manifests as a spatially coherent series of events, transient spikes triggered by local intensity change, distributed along its length. Because event cameras respond asynchronously to brightness gradients, a moving or vibrating DLO produces dense event activity at its edges while the static background remains largely silent. However, the raw event stream is contaminated by sensor noise (isolated hot pixels, transistor shot noise) and, in industrial settings, by periodic flicker from artificial lighting. To suppress these artifacts, the proposed pipeline applies a hardware-accelerated filter with spatio-temporal filtering which suppresses noise by retaining only those events that are corroborated by spatially and temporally neighboring events within a defined radius and time window, effectively discarding isolated, spurious activations. In earlier experiments an Edgelet Tracking Algorithm was additionally evaluated, which fits local oriented edgelets to spatiotemporal event clusters and tracks them across successive time slices; although this provides compact edge-segment descriptors well suited to rigid contour tracking, it was found to be less effective for the highly deformable geometry of DLOs, where the curvature changes continuously along the object. The filtered events are instead accumulated into polarity-separated time surfaces, from which a recentness map and subsequent binary mask extraction yield the per-frame observations fed to the CPD tracker. The morphological refinement of the event mask is adapted to the segmented DLO diameter $d_{seg}$: the closing and opening structuring elements and the number of closing iterations are scaled proportionally to $d_{seg}$, so that gap-bridging and noise removal match the physical thickness of the DLO before the observations $X$ are extracted. 

Kalman Filter assumes Gaussian uncertainty and roughly linear dynamics, which works well for slowly deforming DLOs but struggles with sudden large deformations since it can't represent non-Gaussian distributions. Optical Flow (e.g. Farnebäck) would be the fastest algorithm for Event-based Cameras, but drifts over time and struggles with large deformations and occlusions as it is a pixel-based approach, not a geometrical one. Neither approach maintains an explicit ordered representation of the DLO, which is a prerequisite for downstream manipulation tasks. Therefore, a geodesic-regularized non-rigid registration approach based on the CPD framework is adopted. 

The event-based processing pipeline begins by reading the asynchronous output of the event camera in small temporal batches. After filtering, the remaining ON and OFF events are written into two separate timestamp memories, one for each polarity, which store the most recent event time at every pixel location.

These polarity-specific timestamp memories are then transformed into a recentness map R based on Surface of Active Events (SAE) \cite{lagorce_hots_2017}. The recentness value ${TS}(x, y;\, t)$ at pixel evaluated at the current time $t$ attains high values if recently fired, whereas older activity decays over time. $\tau$, the exponential decay constant, sets how fast events fade. As a result, the recentness map provides a compact spatial summary of the most recent motion in the scene while preserving the temporal responsiveness of the event stream. The map is subsequently thresholded to isolate active DLO regions and is morphologically refined to remove small artifacts and bridge short discontinuities, yielding a binary mask $M_t$ of the moving DLO. 

\begin{equation}
    \mathrm{TS}(x, y;\, t) = \exp\!\left(-\frac{t - T(x,y)}{\tau}\right) 
\end{equation}
Since our application requires only foreground presence rather than edge polarity, both channels are collapsed into a single recentness map $R$
\begin{equation}
   R(x,y) = \max\bigl(\mathrm{TS}^{+},\,\mathrm{TS}^{-}\bigr)
\end{equation}

From $M_t$, an ordered geometric observation of the DLO is extracted. A Principal Component Analysis (PCA)-based centerline is first computed, as it remains robust under local gaps caused by vibration nodes and regions that temporarily produce no events. When sufficient connectivity is present, a skeleton path is additionally extracted via progressive gap-bridging to obtain a more explicit geometric structure. The representation that best captures the full visible extent of the DLO is selected and uniformly resampled to produce a fixed set of ordered observation points $X = \{\mathbf{x}_n\}_{n=1}^{N}$, which serves as input to the tracking stage.

\subsection{CPD Tracker}

\begin{figure*}
    \centering
    \includegraphics[width=1\textwidth]{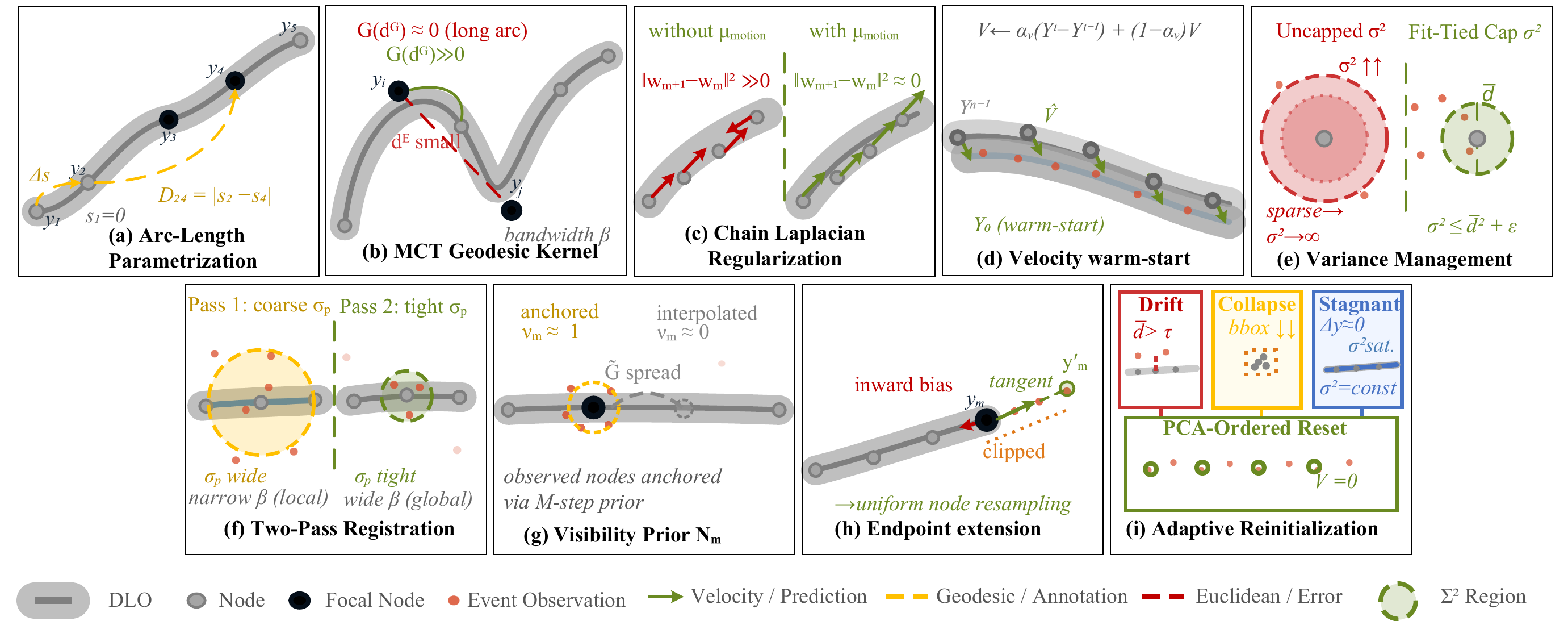}
    \caption{The CPD tracker producing $Y_t$ builds upon established components from prior work, (a) arc-length parametrization and (b) MCT geodesic kernel. To account for DLO velocity $v_{\mathrm{DLO}}$ and improve robustness at higher speeds, (c) chain Laplacian regularization and (d) velocity warm-start are introduced. The 
assumed constant DLO diameter~$d$ is taken into account via component~(e) 
to condition the tracker on event-based input and reduce computational cost. The 
assumed DLO diameter~$d$ is taken into account via component (f) 
to condition the tracker on event-based input and reduce computational cost. Robustness against measurement errors is addressed through the
component (g).Robustness against measurement errors is addressed through 
components (h) and~(i).
}
    \label{fig:formulas_event_tracking}
\end{figure*}

The tracking algorithm with additions as displayed in figure~\ref{fig:formulas_event_tracking} builds on TrackDLO \cite{xiang_trackdlo_2023} and is extended with modifications to improve robustness under sparse and asynchronous event-based observations. 

\paragraph{Arc-length parametrization}
The DLO is represented as an ordered chain of $M$ control nodes $\mathbf{Y} = \{\mathbf{y}_m\}_{m=1}^{M}$ parametrized by arc length,
\begin{equation}
    s_1 = 0, \qquad
    s_i = \sum_{k=2}^{i}\|\mathbf{y}_k - \mathbf{y}_{k-1}\|_2, \quad
    i = 2,\ldots,M,
    \label{eq:arclength}
\end{equation}
so that the geodesic distance between any two nodes is simply $D_{ij} = |s_i - s_j|$. 

\paragraph{MCT geodesic kernel}
Node deformation is regularized by the second-order MCT kernel
\begin{equation}
    G(d) = \frac{1}{4\beta^2}
    \exp\!\left(-\frac{\sqrt{2}\,|d|}{\beta}\right)
    \!\left(2|d| + \sqrt{2}\beta\right),
    \label{eq:kernel}
\end{equation}
where $d = D_{ij}$ and $\beta$ controls the geodesic bandwidth. Using arc length rather than Euclidean distance prevents spatially adjacent but topologically distant segments from coupling, which is essential for DLOs that fold or self-intersect. Observations are assigned to nodes via a two-anchor geodesic E-step, and node positions are updated by minimizing the MCT-regularized objective, as described in~\cite{xiang_trackdlo_2023, myronenko_point_2010}. We adapt the per-node visibility decay to 2D event observations as described below.

While TrackDLO provides a principled geometric foundation for DLO tracking, it is designed for RGB-D sensors that produce dense and spatially uniform point clouds at fixed frame rates. In contrast, event-based sensing produces spatially sparse and temporally asynchronous observations concentrated along intensity gradients. Although TrackDLO leverages motion coherence to propagate information between frames, its formulation assumes sufficiently dense observations at each time step and does not explicitly integrate temporally asynchronous measurements across multiple event batches. Furthermore, it lacks an explicit re-detection or recovery mechanism once data association fails. These limitations become critical when applying the method to event-based data, motivating the following extensions.

\paragraph{Chain Laplacian regularization}
In TrackDLO, the M-step regularizes deformation magnitude through the term $\lambda \sigma^2 \mathbf{I}$ and enforces spatial smoothness via the kernel matrix $\mathbf{G}$. This smoothness is defined with respect to pairwise geodesic distances along the curve, but does not explicitly penalize differences between adjacent deformation vectors along the chain. As a consequence, neighboring nodes along the DLO may undergo inconsistent local deformations when observations are sparse or unevenly distributed, as is typical in event-based sensing. To address this limitation, the M-step is augmented with a chain Laplacian regularization term,
\begin{equation}
    \mathbf{A} \;\leftarrow\; \mathbf{A} + \mu_{\text{motion}} \mathbf{L}^\top \mathbf{L},
\end{equation}
where $\mathbf{L} \in \mathbb{R}^{(M-1)\times M}$ is the first-difference operator along the node ordering, such that
$(\mathbf{L}\mathbf{W})_m = \mathbf{w}_{m+1} - \mathbf{w}_m$. This modification corresponds to augmenting the M-step objective with an
additional quadratic penalty of the form
\begin{equation}
    \mu_{\text{motion}} \sum_{m=1}^{M-1}
    \left\| \mathbf{w}_{m+1} - \mathbf{w}_m \right\|^2,
\end{equation}
which enforces local consistency of the deformation field along the DLO centerline.

While the kernel $\mathbf{G}$ promotes global smoothness based on spatial proximity, the proposed term explicitly enforces topology-aware regularization aligned with the underlying one-dimensional structure of
the object.

\paragraph{Inter-frame velocity warm-start}
TrackDLO initializes each registration from $\mathbf{Y}_0 = \mathbf{Y}^{t-1}$, effectively assuming negligible inter-frame motion. At event-camera rates, the DLO may move beyond the region where the Gaussian correspondence
weights remain significant between successive batches, leading to degraded
data association.

To address this, we maintain a per-node velocity field
$\mathbf{V} \in \mathbb{R}^{M \times 2}$, updated after each registration
using an exponential moving average of the observed node displacement,
\begin{equation}
    \mathbf{V} \leftarrow
    \alpha_v \bigl(\mathbf{Y}^t - \mathbf{Y}^{t-1}\bigr)
    + (1 - \alpha_v)\,\mathbf{V},
    \label{eq:ema}
\end{equation}
where $\alpha_v \in (0,1)$ is a smoothing parameter that controls the
relative influence of the current displacement and the accumulated velocity estimate, $\mathbf{Y}^{t-1}$ and $\mathbf{Y}^t$ denote the node positions before and after registration at time $t$, respectively. This provides a
temporally smoothed estimate of node motion and reduces sensitivity to
frame-to-frame noise.

Prior to each new registration, the velocity field is propagated through
the geodesic kernel to obtain a motion-consistent warm start,
\begin{equation}
    \mathbf{Y}_0 = \mathbf{Y}^{t-1} + \tilde{\mathbf{G}}\,\mathbf{V},
    \qquad
    \tilde{G}_{ij} = \frac{G_{ij}}{\sum_k G_{ik}},
    \label{eq:warmstart}
\end{equation}
where $\tilde{\mathbf{G}}$ is the row-normalized kernel matrix.

Propagating the velocity through $\tilde{\mathbf{G}}$, rather than
applying it independently per node, enforces spatial coherence in the
predicted motion. In this formulation, each node's displacement is a
geodesically weighted average of neighboring velocities, allowing
nodes with sparse or missing observations to receive motion information
from nearby regions of the curve.

\paragraph{Noise variance management}
In event-based settings, where observations are spatially sparse and temporally 
fragmented, reliable estimation of $\sigma^2$ requires measures as independent 
re-estimation at each registration step risks a degeneracy in which limited data 
inflates $\sigma^2$, weakening data association. To prevent this feedback instability, 
$\sigma^2$ is therefore bounded from above by a fixed threshold $\sigma^2_{\max}$.

This estimation is addressed with two mechanisms. At initialization or after recovery, $\sigma^2$ is set proportional to the node spacing,
\begin{equation}
    \sigma^2_0 = \left(\frac{\bar{s}}{2}\right)^{\!2},
\end{equation}
where $\bar{s}$ is the mean inter-node arc-length spacing, reflecting
the assumption that observations lie within the neighborhood of their
nearest node. During tracking, $\sigma^2$ is constrained by a fit-tied cap,
\begin{equation}
    \sigma^2 \leftarrow \min   \! \bigl(\sigma^2,\;\bar{d}^{\,2} + \varepsilon\bigr),
\end{equation}
where $\bar{d}$ is the mean node-to-observation residual and $\varepsilon = 20\text{px}^2$ is a margin that permits $\sigma^2$ to exceed the current fit under fast motion. This breaks the runaway feedback while still permitting $\sigma^2$ to grow when large deformations require a broader correspondence support.

\paragraph{Two-pass registration with soft observation weighting}
Hard spatial pruning introduces a discontinuity in the data association process: observations beyond a fixed distance are discarded entirely,which can cause failure when the DLO undergoes rapid motion. To reduce this effect, the strict per-node threshold is relaxed into a coarse trust region that only removes observations far from the curve, while the fine-grained association is governed by a predominantly soft weighting scheme combined with a two-pass registration strategy.

Each observation is assigned a Gaussian weight based on its distance to the predicted node configuration,
\begin{equation}
    w_n = \exp\!\left(-\frac{d_n^2}{2\sigma_p^2}\right),
\end{equation}
where $d_n$ denotes the minimum distance from observation $\mathbf{x}_n$
to the node set and $\sigma_p$ controls the weighting scale. A coarse
pre-processing pass uses a relaxed scale to retain distant but
potentially valid observations and obtain a robust initial alignment.
In a subsequent refinement pass, weights are recomputed using the
updated node positions with a tighter scale. Beyond the observation weighting scale, the two passes also differ in their regularization regime: the pre-processing pass uses a narrow kernel bandwidth so that each node moves primarily toward its nearest observations, while the main pass uses a wide bandwidth that couples nodes over a larger geodesic radius and enforces global shape smoothness.

These weights are incorporated into the E-step by scaling each
observation prior to normalization, yielding

\begin{equation}
p(m \mid \mathbf{x}_n) =
\frac{
    w_n\,\nu_m \exp\!\left(-\frac{(d^G_{mn})^2}{2\sigma^2}\right)
}{
    \sum_{k=1}^{M}
    w_n\,\nu_k \exp\!\left(-\frac{(d^G_{kn})^2}{2\sigma^2}\right) + c
}
\end{equation}

where $w_n$ is an observation-dependent weight, $\nu_m$ is a per-node visibility term evaluated at the current node estimate at each EM iteration, and $c$ is the outlier constant of the Gaussian mixture,
\begin{equation}
    c = (2\pi\sigma^2)^{D/2}\,\frac{\mu}{1-\mu}\,\frac{M}{N},
\end{equation}
with $D$ the spatial dimension, $\mu \in (0,1)$ the assumed outlier ratio,$M$ the number of nodes, and $N$ the number of observations. Because $w_n$ multiplies only the data terms and not the outlier constant $c$,observations with low weight shift toward the outlier distribution rather than being removed abruptly. This formulation departs from the standard CPD likelihood and is treated as a regularized registration scheme.

\paragraph{Visibility-aware correspondence prior}
To stabilize the registration under uneven observation coverage, we
introduce a per-node visibility weight computed from the pre-aligned
configuration,
\begin{equation}
\nu_m^{\text{pre}} = \exp\!\left(-k_{\text{vis}}\,d_m^{\min}\right),
\end{equation}
where $d_m^{\min}$ is the distance from node $m$ in $\mathbf{Y}_{\text{pre}}$ to its nearest observation. Well-supported nodes (small $d_m^{\min}$, high $\nu_m^{\text{pre}}$) are strongly anchored to their positions in $\mathbf{Y}_{\text{pre}}$ through a correspondence prior in the M-step, preventing the main CPD pass from over-correcting regions that the pre-processing step has already aligned. In contrast, poorly-supported nodes receive a weaker prior, allowing the deformation field to interpolate smoothly from neighboring well-observed regions through the kernel $\mathbf{G}$.

\paragraph{Endpoint extension}
Due to the symmetric formulation of the CPD objective, nodes near the
endpoints receive observations from only one side, leading to a
systematic inward bias. This effect is exacerbated in event-based data,
where observations near the tips are typically sparse.

To correct this, both endpoints are extended after each registration
step along their local outward tangent direction. Candidate observations
are selected from the full, unweighted observation set based on their
alignment with the tangent direction and proximity to the curve axis.
The farthest valid observation determines the extension length, which is
clipped to prevent excessive extrapolation. The curve is subsequently
resampled to maintain uniform node spacing. This ensures that endpoints follow the true extent of the DLO even under sparse or uneven observation coverage.

\paragraph{Adaptive re-initialization}
An explicit mechanism for recovering from tracking failure is added with three complementary triggers. A \emph{drift-threshold} trigger fires when the mean node-to-observation residual $\bar{d}$ exceeds a limit, indicating nodes have lost contact with the DLO. A \emph{stuck-span} trigger fires when the bounding-box diagonal of $\mathbf{Y}^t$ remains below a minimum extent across several consecutive frames, indicating node
collapse onto a noise cluster. The third trigger detects stagnation when the noise variance remains saturated while node motion is minimal, indicating that the tracker is no longer responding to observations despite not exceeding the drift threshold. All three triggers reinitialize from the ordered centerline path when available, falling back to a PCA-ordered cloud of current observations and reset $\mathbf{V} = \mathbf{0}$, preventing accumulated motion
from pulling the freshly placed nodes to the lost position. The noise variance is reset using the cold-start estimate introduced in the noise-variance management above.

\subsection{Fusion Mechanism}

\paragraph{Time synchronization}
\begin{figure}
    \centering
    \includegraphics[width=1\linewidth]{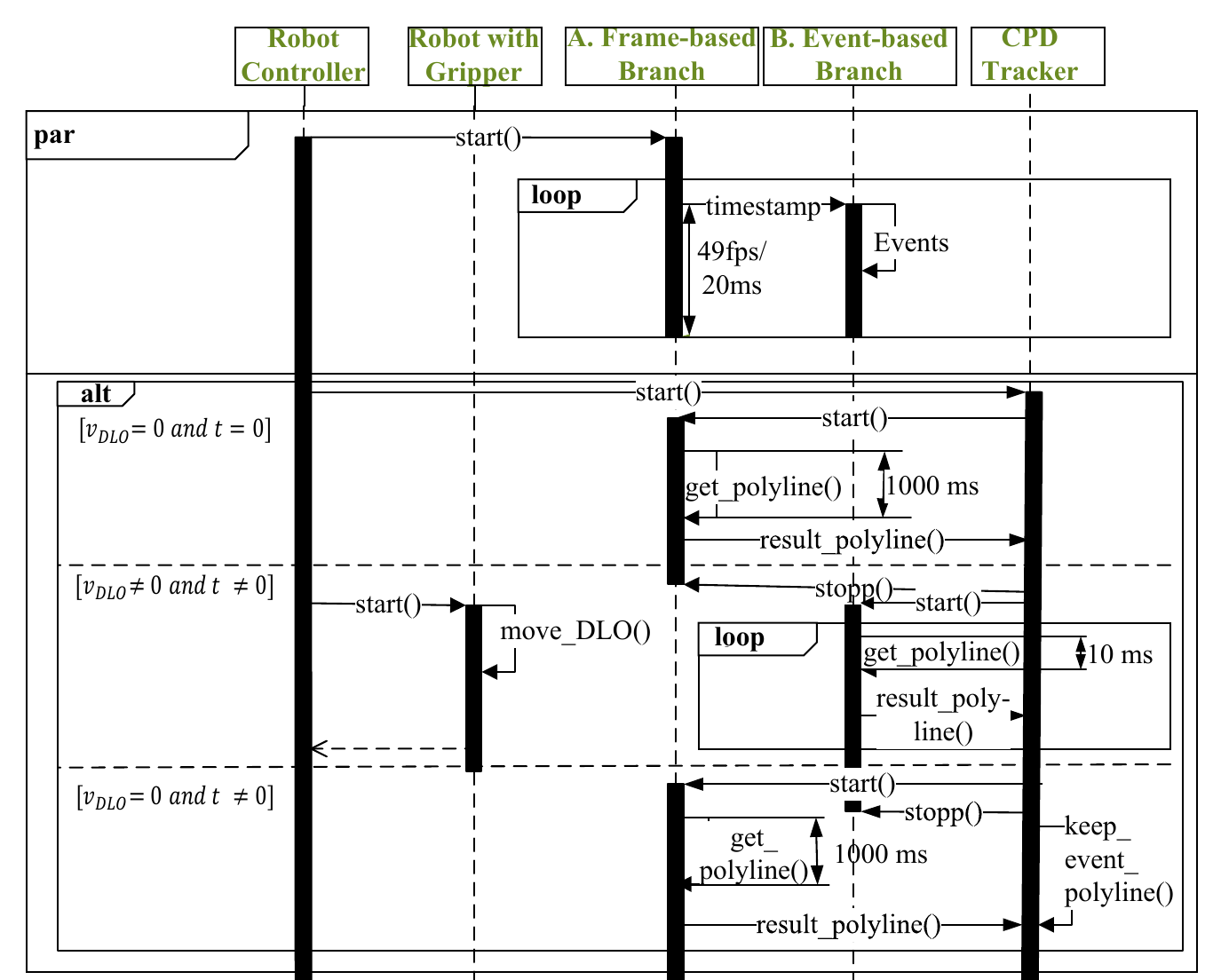}
    \caption{Sequence diagram of the respective timing between the different instances}
    \label{fig:SequenceDiagram}
\end{figure}

In order to generate minimal delays, hardware synchronization is chosen as presented in figure~\ref{fig:SequenceDiagram}. The image sensor receives a signal from the robot controller at the beginning of each sequence. The image sensor and event sensor are synchronized via hardware, using an external computer that delivers trigger signals to both sensors at the same time. Synchronized recordings were obtained by routing the RGB camera's post-exposure trigger signal directly to the event camera's external trigger input, as described above. A manifest file was produced at recording time, associating each saved RGB frame with its corresponding trigger timestamp in the event clock domain, thereby establishing an unambiguous temporal correspondence between the two modalities. 

\paragraph{Spatial Calibration using Cross-Modal Homography}

The homography calibration is performed once per camera mounting configuration. For each correspondence pair, the pixel coordinate in the RGB image is recorded first, followed by the corresponding pixel coordinate in an accumulated event frame generated from the first seconds of the recording. The homography is computed via the Direct Linear Transform (DLT)~\cite{abdel-aziz_direct_2015} within a RANSAC~\cite{fischler_random_1981} robust estimation loop with a reprojection-error of 0.79 px. RANSAC iteratively samples minimal four-point subsets, estimates a candidate homography, and scores it by counting correspondences whose symmetric transfer error falls within the threshold; the hypothesis with the highest inlier count is retained and refined over all its inliers. 

\paragraph{Hibernation and Re-initialization}

While the event-based tracker provides high-rate continuous state estimates and the frame-based segmentation provides geometrically complete but infrequent segmentation, neither is sufficient in isolation: the event-based tracker does not detect any events while the DLO is not moving and requires manual tuning, while SAM 3 alone cannot sustain real-time tracking at the required temporal resolution. 
At each event batch, the fraction of active mask pixels is computed:
\begin{equation}
  \rho_t = \frac{|M^{\mathrm{ev}}_t|}{H \cdot W}
\end{equation}
The DLO is considered \emph{static} if $\rho_t < \rho_{\mathrm{thr}}$, where $\rho_{\mathrm{thr}}$ is the threshold below which the DLO is assumed to be static. In this case, hibernation is leveraged by keeping the last generated polyline from the event-based tracker and a re-initialization of the SAM 3 segmentation is started, to achieve more accurate results. This SAM 3 segmented polyline is then again used for reinitializing the event-based tracker when movement of the DLO is detected.

\section{Experimental Evaluation}

\subsection{Evaluation Procedure and Metrics}

The evaluation focuses on performance under industrially relevant manipulation conditions with emphasis on four critical dimensions derived from the problem statement \ref{Problem Statement}: robustness to industrial movement speeds of up to 2000 mm/s, identity preservation for each DLO identity $i \in \{1,\dots,K\}$, segmentation and geometric accuracy for the estimate $\hat{Y}_t^{(i)}$ with \ref{eq:estimation_ground_truth} minimized, and real-time suitability for closed-loop automation. For a real-time suitability the $\Delta t$ between each DLO estimation $\hat{Y}_t^{(i)}$ needs to be at 10 ms in order to support DLO movement at robot speeds of 2000 mm/s. Unlike in \cite{caporali_rtdlo_2023, caporali_fastdlo_2022} and others which prioritize segmentation accuracy or inference speed in isolation, the proposed evaluation incorporates temporal consistency as a criterion. Accordingly, both per-frame accuracy and cross-frame identity preservation are considered jointly in the comparative analysis.
The tracking accuracy is quantified by the average per-node error, defined as
\begin{equation}
  e_t = \frac{1}{M} \sum_{m=1}^{M}
        \left\| \mathbf{y}_m^t - \hat{\mathbf{y}}_m^t \right\|_2 ,
  \label{eq:per_node_error}
\end{equation}
where $M$ denotes the number of nodes, $\mathbf{y}_m^t$ is the ground-truth
position of node~$m$ at time step~$t$, and $\hat{\mathbf{y}}_m^t$ is the
corresponding estimated position. The overall tracking error is reported as
the temporal mean
\begin{equation}
  \bar{e} = \frac{1}{T} \sum_{t=1}^{T} e_t ,
  \label{eq:mean_error}
\end{equation}
averaged over all $T$~frames of a sequence. However, as no frame representation anymore is present, this is replaced by T update step. 
 
In addition to the per-node correspondence error, a point-to-curve metric is used to account for tangential drift of estimated nodes along the DLO, a scenario in which index-based pairing can overestimate the true geometric error. Following \cite{xiang_trackdlo_2023}, let $\mathrm{PWL}(\mathbf{Y}^t)$ denote the piecewise-linear curve induced by the ordered ground-truth node sequence $\mathbf{Y}^t$, comprising both node positions and the linear segments between adjacent nodes. The distance from an estimated node $\hat{\mathbf{y}}_m^t$ to the ground-truth curve is defined as the minimum Euclidean distance to any 
point on this piecewise-linear structure 
\begin{equation}
    d(\hat{\mathbf{y}}_m^t, \mathrm{PWL}(\mathbf{Y}^t))
    = \inf_{\mathbf{p} \in \mathrm{PWL}(\mathbf{Y}^t)}
      \|\hat{\mathbf{y}}_m^t - \mathbf{p}\|_2 .
    \label{eq:ptc_dist}
\end{equation}

and the per-step point-to-curve error is adapted from 
\cite{xiang_trackdlo_2023}

\begin{equation}
    e_t^{\mathrm{ptc}}
    = \frac{1}{M}\sum_{m=1}^{M}
      d(\hat{\mathbf{y}}_m^t, \mathrm{PWL}(\mathbf{Y}^t)) ,
    \label{eq:ptc_error}
\end{equation}

with the overall point-to-curve error reported as the temporal mean

\begin{equation}
    \bar{e}^{\mathrm{ptc}} 
    = \frac{1}{T}\sum_{t=1}^{T} e_t^{\mathrm{ptc}} ,
    \label{eq:mean_ptc_error}
\end{equation}

A solely inference time based comparison is not sufficient, as it has the shortcomings that an algorithm suffering from ID switch outperforms one without this drawback. ID switch is tracked by whether the number of the identified DLO changes and therefore is no longer trackable in dense environments.

The proposed method, MotionDLO, is benchmarked against three state-of-the-art approaches: TrackDLO, RT-DLO, and SAM 3 \cite{xiang_trackdlo_2023,caporali_rtdlo_2023} under controlled and repeatable experimental conditions. UniStateDLO \cite{lv_unistatedlo_2025} could not be benchmarked as the code was not publicly released. The benchmark targets the most recent state-of-the-art representative of each algorithmic family: TrackDLO supersedes CDCPD, CDCPD2, and GLTP, while RT-DLO supersedes FastDLO, Ariadne, and Ariadne+, making comparison against predecessors implicit. Segmentation baselines are omitted, as SAM has been demonstrated to achieve superior segmentation accuracy over prior methods. DLOFTBs builds upon FastDLO, yet it uses B-Splines. B-Splines break when occlusions occur, making the approach unsuitable for the given boundaries as several DLOs are in one image.

\subsection{Experiment Setup and Dataset Generation}

\begin{figure}
    \centering
    \includegraphics[width=1\linewidth]{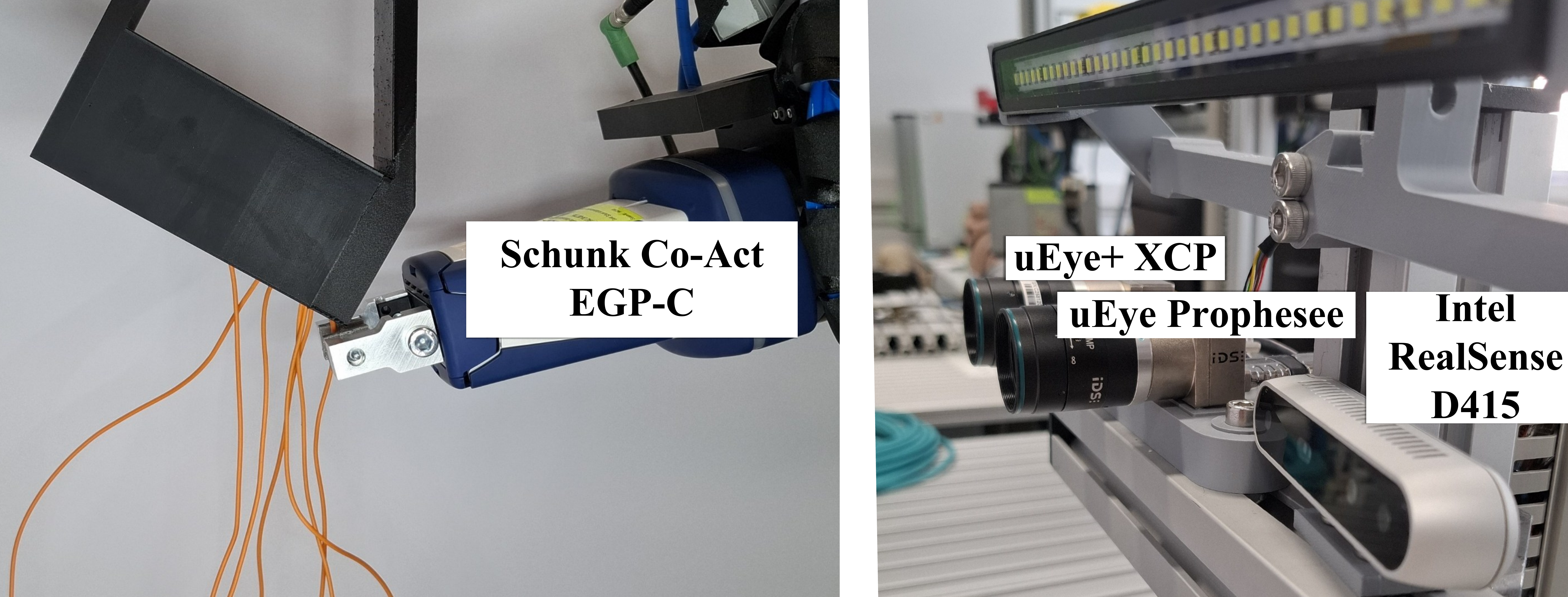}
    \caption{A single DLO is grasped in an entangled scenario and manipulated by the robot gripper (left). An event-based camera, a frame-based camera, and a depth sensor are mounted in a fixed configuration to jointly observe the DLO motion trajectory (right).}
    \label{fig:ExperimentalSetup}
\end{figure}

As shown in Figure \ref{fig:ExperimentalSetup} the experimental setup consists of three camera systems to enable comparison across all algorithms. The RGB-D system employs an Intel RealSense D415 camera with 1280 × 720 resolution at 30 fps. The D415 variant is selected due to its improved depth accuracy for thin DLOs. The RGB system uses an IDS uEye+ U3-3680XCP color camera based on the Onsemi CMOS sensor, providing 2592 × 1944 pixel resolution at 49 fps and equipped with a 6.0 mm C-mount lens. The global start shutter sensor eliminates motion artifacts during fast DLO manipulation, while high-resolution imaging enables sub-millimeter DLO boundary localization critical for thin DLO detection. \cite{caporali_deformable_2023} The event camera system employs a Sony Prophesee Sensor IMX636 operating at 1280 x 720, co-located with the other cameras and equipped with a 6.0 mm C-mount lens.  

All cameras were mounted at a working distance of 200 mm from the manipulation workspace, with optical axes aligned parallel to the scene plane to maintain consistent geometric conditions throughout the study. DLO manipulation was executed using a robotic gripper performing repeatable trajectories to ensure experimental consistency, and illumination was provided by LED panels to maintain controlled lighting conditions across trials. A Fanuc CRX 10 iA$/$ L was used for DLO movement equipped with a Schunk Co-act EGP-C collaborative gripper. All experiments were conducted on a workstation equipped with an NVIDIA RTX 6000 Ada Generation GPU with CUDA 13.0 support. Software environments were configured independently for each algorithm to ensure compatibility and stable execution.

While temporal alignment is established through the hardware trigger mechanism described above, the two sensors also differ in their spatial coordinate frames: the RGB camera operates at $2592 \times 1944$ pixels, whereas the event sensor operates at $1280 \times 720$ pixels, and the two optical axes are not perfectly co-located due to the physical separation of the camera housings. 

Both cameras were first calibrated intrinsically using the Prophesee Metavision calibration toolbox \cite{metavision_calibration}, which is compatible with both sensors as they are IDS-based devices supported by the Prophesee SDK. A standard checkerboard pattern was used to estimate the camera matrix and lens distortion coefficients for each sensor independently. To project segmentation masks produced in RGB image space into event sensor coordinates, a planar projective homography $\mathbf{H} \in \mathbb{R}^{3 \times 3}$ is additionally estimated directly between the two image planes \cite{hartley_multiple_2003}.
 
\begin{figure}
    \centering
    \includegraphics[width=0.75\linewidth]{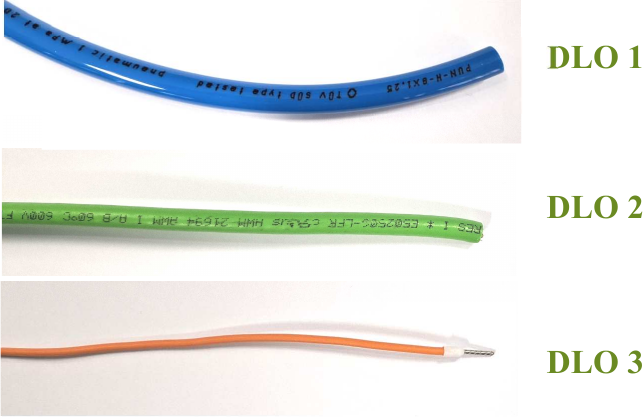}
    \caption{DLO~1 is a standard pneumatic plastic tube with a reflective surface and a hollow core. DLO~2 is a PROFINET cable, while DLO~3 is a switchgear cabinet 
wire with a copper core of 1.5\,mm diameter.}
    \label{fig:DLOTypes}
\end{figure}

Building upon this controlled acquisition setup, the benchmark is organized along three orthogonal challenge axes designed to systematically stress DLO perception under realistic manipulation conditions. These axes include manipulation speed, DLO thickness, and multi DLO interaction.As shown in figure \ref{fig:DLOTypes} three types of DLOs from different domains are chosen for a wide variety. DLO thickness is varied to explicitly test sensitivity to thin structures, which are known to challenge both depth-based and RGB segmentation methods. DLO 1 is a plastic tubing with an outer diameter of 6 mm, reflecting TPE-U(PU) material, a minimal bending radius of 21 mm and 0.0287 kg/m. DLO 2 is a PROFINET type B, 4-core, shielded, CAT 5e cable with an outer diameter of 6.5 mm, a non-reflective PVC outer sheath, a minimal bending radius of 32.5 mm and 0.068 kg/m. DLO 3 is a single core cable with an outer diameter of 4 mm and a minimal bending radius of 13.6 mm, a non-reflective PVC outer sheath and 0.021 kg/m. DLO 1 is used for air pressure transmission, DLO 2 for Profinet communication and DLO 3 for e.g., power transmission in switch cabinets.

For evaluation purposes two datasets are acquired. Both datasets consist of measurement sequences with 1,858 images, 6.2 GB of event data and a .json file assigning the respective frames to event-based timestamps from three industrially used DLO types measured at three different speed levels acquired with an event and a frame-based camera simultaneously. A measurement sequence duration is defined by the speed of the robotic gripper which grips one DLO and moves it through the field of view (FOV). The first dataset contains in each scenario four overlapping DLOs from which one is gripped and moved with a robotic gripper. It is used as an evaluation dataset for benchmarking MotionDLO against the Event-based branch of Motion DLO as well as the SAM 3 branch each as a stand-alone solution and benchmarking against single-frame tracking methods such as RT-DLO. The dataset is publicly available \cite{hartmann_event-based_2026}.

Current cross-frame tracking does not work for this scenario at all as TrackDLO does only work for single DLOs in the field of view. Therefore, a second dataset is acquired using a depth sensor, with measurements captured for all three DLO types. However, TrackDLO only works for DLO types 1 and 2 as the DLO diameter of DLO type 3 with 4 mm is too thin for TrackDLO to detect. The DLO is moved by a robot which is accelerated from the gripping position at an average acceleration in a linear motion until it reaches a robot speed of 50$\%$ , 75$\%$ and 100$\%$ respectively. Table~\ref{tab:robot_speed} displays the resulting DLO speed based upon this acceleration. The maximum speed is measured over the last 30 update steps, while the average is the median of the whole measurement. 

\begin{table}[h]
\caption{TCP speed measurements at different robot speed settings.}
\label{tab:robot_speed}
\centering
\renewcommand{\arraystretch}{1.35} 
\setlength{\tabcolsep}{9pt}        

\begin{tabular}{|c|c|c|c|}
\hline
\textbf{Robot Speed (\%)} & \textbf{Measure} & \textbf{px/s} & \textbf{mm/s} \\
\hline
\multirow{2}{*}{50}  
    & average & 1398 & 893.7  \\
\cline{2-4}
    & max     & 2339 & 1495.0 \\
\hline
\multirow{2}{*}{75}  
    & average & 1923 & 1229.0 \\
\cline{2-4}
    & max     & 2363 & 1510.3 \\
\hline
\multirow{2}{*}{100} 
    & average & 2312 & 1477.7 \\
\cline{2-4}
    & max     & 3529 & 2255.1 \\
\hline
\end{tabular}

\vspace{6pt}

\end{table}
\subsection{Experiment Results}

\subsubsection{Identity Preservation in Multi-DLO Scenes}  

\begin{figure}
    \centering
    \includegraphics[width=1\linewidth]{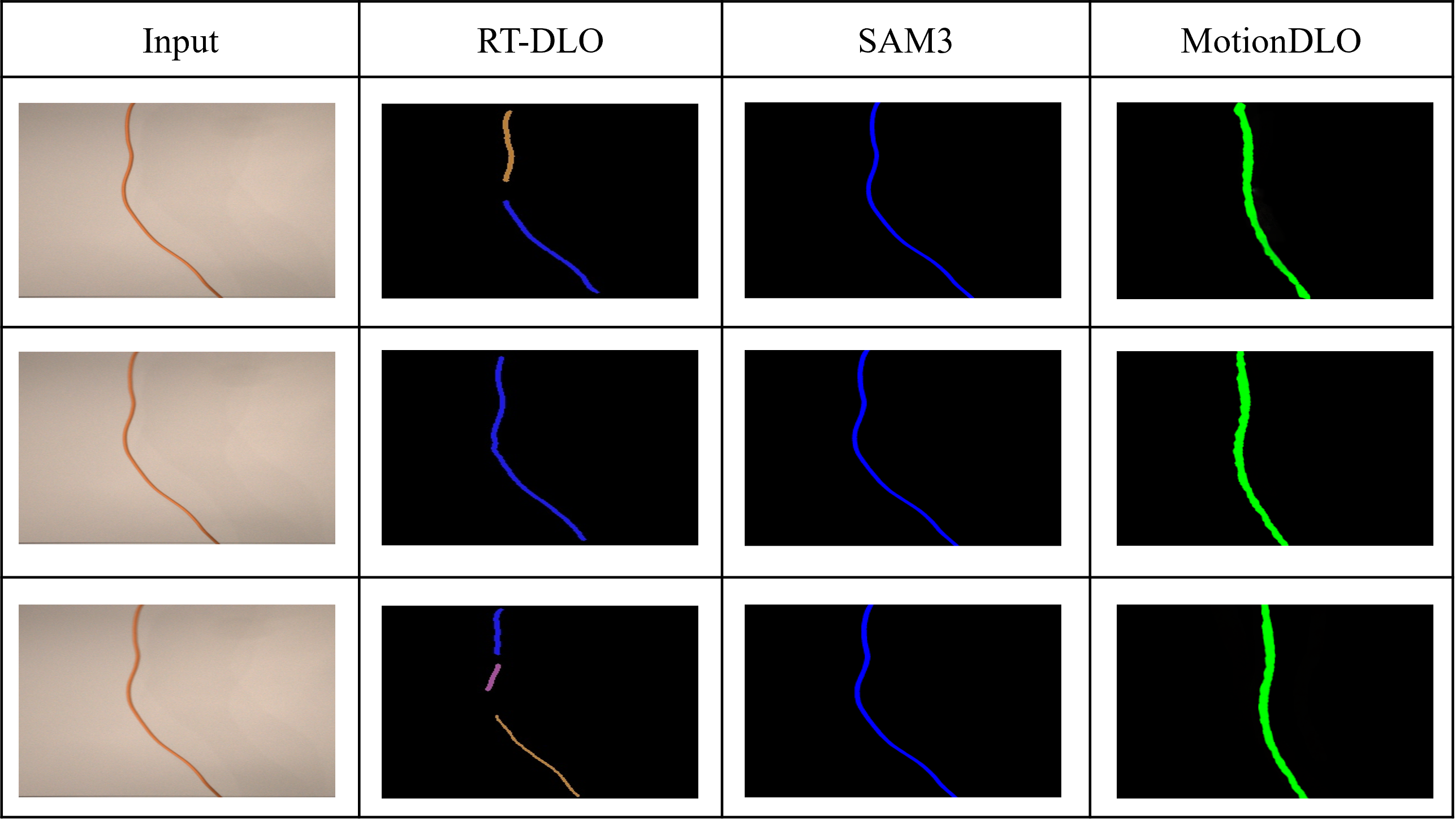}
    \caption{Qualitative evaluation of the observed ID switch. The input image shows the RGB image at three consecutive timestamps, while RT-DLO, SAM 3 and MotionDLO show the respective results.}
    \label{fig:ID switch}
\end{figure}

The ID switch behavior is evaluated on both datasets, both quantitatively and
qualitatively. RT-DLO, SAM~3, and MotionDLO were each executed independently on the 
same single-cable sequences from Dataset~1, and representative frames 
were extracted to produce the qualitative comparison in figure~\ref{fig:ID switch}. 
For the first 20 update timestamps of each sequence, the number of DLO instances and 
the corresponding ID switches were determined by manual annotation. Switches are 
counted from detection and association errors, namely instances that cease to be 
detected, newly introduced instances, and changes in instance assignment across 
consecutive frames (Table~\ref{tab:ID-switches1}).

The qualitative window is shorter than the quantitative one for two reasons. First, 
the orientation CNN of RT-DLO operates on a fixed $15\times15$ patch centered at each skeleton vertex~\cite{caporali_rtdlo_2023}, which requires every candidate vertex to lie at least $7$\,px from the image boundary. In the sequences evaluated here, the segmentation mask contained valid foreground pixels and a non-empty skeleton in the majority of frames, yet all skeleton candidates fell within this boundary margin, so graph generation produced no output for those frames. Second, beyond update step~11 the RT-DLO and SAM~3 segmentations break down, as the manipulation at robot speed~75 $\%$ induces sufficient motion blur to suppress reliable foreground extraction despite additional lights, strong gain, an open aperture and low exposure times. The qualitative comparison is therefore restricted to the contiguous range in which all three methods yield a valid output. While the MotionDLO mask is less geometrically precise than the one derived from SAM~3, this is traded for a substantially lower inference cost. The depicted sequence corresponds to DLO~3 at robot speed~75 $\%$ and is chosen as DLO 3 is the most deformable and thinnest of the dataset. The robot speed 75 $\%$ was chosen as robot speed 100 $\%$ delivers less images without motion blur. 

Within this window, RT-DLO splits the single DLO into several distinct instances 
across consecutive frames, constituting an identity switch. SAM~3 produces a 
geometrically consistent segmentation in the single-cable scenario without switches; 
in multi-cable scenes, however, it is observed to lose track of the moving cable, 
as it provides no mechanism for associating instances across frames~\cite{caporali_rtdlo_2023}. MotionDLO, in contrast, maintains a continuous, identity-stable estimate throughout the sequence via the event-based branch.

\begin{table}[t]
    \centering
    \caption{Count of ID switches between consecutive timestamps for the DLO 3 at robot speed 75 $\%$ sequence. Ideal result for the single DLO scenario is no ID switch and one detected DLO.}
    \label{tab:ID-switches1}
    \scriptsize
    \renewcommand{\arraystretch}{1.15}
    \setlength{\tabcolsep}{4pt}

    \begin{tabular}{@{}c cc cc cc@{}}
        \toprule
        & \multicolumn{2}{c}{RT-DLO}
        & \multicolumn{2}{c}{SAM 3}
        & \multicolumn{2}{c}{MotionDLO} \\
        \cmidrule(lr){2-3}
        \cmidrule(lr){4-5}
        \cmidrule(lr){6-7}
        Time
        & ID switch & Instances
        & ID switch & Instances
        & Switch & Instances \\
        \midrule
        1  & 0 & 2 & 0 & 1 & 0 & 1 \\
        2  & 1 & 2 & 0 & 1 & 0 & 1 \\
        3  & 2 & 2 & 0 & 1 & 0 & 1 \\
        4  & 2 & 2 & 0 & 1 & 0 & 1 \\
        5  & 2 & 2 & 0 & 1 & 0 & 1 \\
        6  & 1 & 1 & 0 & 1 & 0 & 1 \\
        7  & 1 & 2 & 0 & 1 & 0 & 1 \\
        8  & 1 & 1 & 0 & 1 & 0 & 1 \\
        9  & 1 & 1 & 0 & 1 & 0 & 1 \\
        10 & 1 & 1 & 0 & 1 & 0 & 1 \\
        11 & 1 & 0 & 0 & 1 & 0 & 1 \\
        12 & - & - & - & - & 0 & 1 \\
        13 & - & - & - & - & 0 & 1 \\
        14 & 0 & 0 & - & - & 0 & 1 \\
        15 & - & - & - & - & 0 & 1 \\
        16 & - & - & - & - & 0 & 1 \\
        17 & 1 & 1 & - & - & 0 & 1 \\
        18 & - & - & - & - & 0 & 1 \\
        19 & - & - & - & - & 0 & 1 \\
        20 & - & - & - & - & 0 & 1 \\
        \bottomrule
    \end{tabular}

    \vspace{1mm}
    
\end{table}

\begin{figure}
    \centering
    \includegraphics[width=1\linewidth]{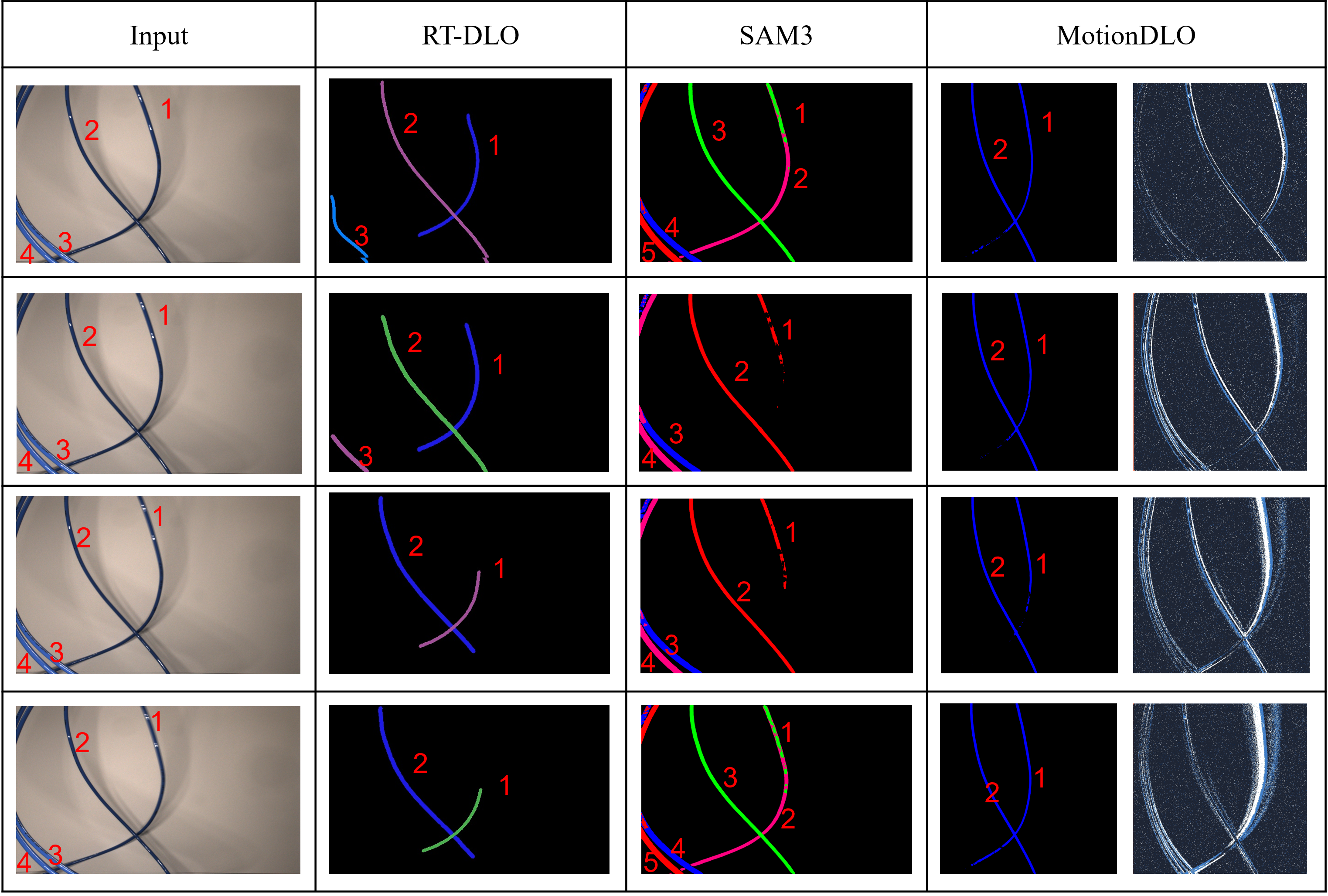}
    \caption{Segmentation results for the multiple wire scenario, where the rightmost DLO is moved and needs to be tracked.}
    \label{fig:MultiDLOSeg}
\end{figure}

For the multi-cable scenario, DLO~1 at robot speed~75 $\%$ was selected for the qualitative and quantitative
evaluation; the corresponding comparison is shown in figure~\ref{fig:MultiDLOSeg}. 
RT-DLO, lacking an inter-frame association mechanism, is unable to distinguish 
the manipulated cable from the surrounding static DLOs and reports the bundle 
as one or several misassociated instances whose boundaries fluctuate across 
consecutive frames. DLO 1 is used instead of DLO 3 for qualitative visualization purposes as the DLOs are more evenly distributed across the images due to the deformable behavior. SAM~3 exhibits an analogous failure mode: the zero-shot segmentation captures the bundled foreground geometry correctly but cannot 
single out the cable currently under manipulation, since no temporal cue is 
available to the per-frame model. MotionDLO, in contrast, exploits the inherent 
motion selectivity of the event stream: only the moving DLO triggers events, so 
the event-based branch isolates the manipulated cable from the static 
surroundings and maintains a stable instance identity over the active 
manipulation phase. The corresponding ID switch counts are reported in Table~\ref{tab:ID-switches-multiple}. As MotionDLO tracks by design only one moving wire as assumed to be relevant for robotic manipulation, no distinction is possible if several wires are manipulated by the robot at the same time. Therefore, if several wires are moved at similar speeds, MotionDLO can lead produce failure causes.

\begin{table}[t]
    \centering
     \caption{Count of ID switches between consecutive timestamps. Ideal result for the multiple DLO scenario is no ID switch and detection of the moving DLO, depending on the algorithm.}
    \label{tab:ID-switches-multiple}
    \scriptsize
    \renewcommand{\arraystretch}{1.15}
    \setlength{\tabcolsep}{4pt}

    \begin{tabular}{@{}c cc cc cc@{}}
        \toprule
        & \multicolumn{2}{c}{RT-DLO}
        & \multicolumn{2}{c}{SAM 3}
        & \multicolumn{2}{c}{MotionDLO} \\
        \cmidrule(lr){2-3}
        \cmidrule(lr){4-5}
        \cmidrule(lr){6-7}
        Time
        & ID switch & Instances
        & ID switch & Instances
        & Switch & Instances \\
        \midrule
        1  & 0 & 2 & 0 & 3& 0 & 1 \\
        2  & 2 & 4 & 0 & 3& 0 & 1 \\
        3  & 2 & 2 & 0 & 3& 0 & 1 \\
        4  & 3 & 3 & 0 & 3& 0 & 1 \\
        5  & 2 & 3 & 0 & 3& 0 & 1 \\
        6  & 3 & 2 & 2& 4& 0 & 1 \\
        7  & 1 & 2 & 0 & 4& 0 & 1 \\
        8  & 1 & 2 & 3& 3& 0 & 1 \\
        9  & 0 & 2 & 4& 4& 0 & 1 \\
        10 & 4 & 4 & 3& 4& 0 & 1 \\
        11 & 4 & 1 & 0 & 4& 0 & 1 \\
        12 & 1 & 1 & 1& 3& 0 & 1 \\
        13 & 0 & 1 & 0& 3& 0 & 1 \\
        14 & 1 & 1 & 3& 2& 0 & 1 \\
        15 & 1 & 1 & 2& 2& 0 & 1 \\
        16 & 2 & 2 & 3& 3& 0 & 1 \\
        17 & 1 & 2 & 3& 1& 0 & 1 \\
        18 & 1 & 2 & 3& 3& 0 & 1 \\
        19 & 0 & 2 & 4& 4& 0 & 1 \\
        20 & 1 & 3 & 2& 4& 0 & 1 \\
        \bottomrule
    \end{tabular}

    \vspace{1mm}
   
\end{table}

\subsubsection{Inference Time and Real-Time Suitability}

For inference measurements, MotionDLO is further split into the substeps recentness map, region of interest masking, mask extraction, candidate selection, skeletonization, CPD and the complete process. The times are calculated as averages over the complete dataset and are reported in Table \ref{tab:inferencetime}. Benchmarking against state-of-the-art algorithms is not possible in a complete scenario as RT-DLO focuses on the mask-extraction, while TrackDLO focuses on the CPD tracking. Therefore, integrating both into a complete pipeline would skew the results and the comparison in the substeps is conducted instead. 

TrackDLO inference time is measured on the green-cable sequence (Dataset 1, robot speed 50 $\%$) over the 169 tracking steps preceding tracking failure. Skeletonization-based initialization requires 669 ms and the per-step CPD registration averages 15.22 ms, the EM optimization converges in a few milliseconds when inter-frame displacement is small, yet requires substantially more iterations as motion grows. The focus of  TrackDLO is not the skeletonization as it takes 669 ms, yet it is taken into account to be able to compare. As TrackDLO's correspondence estimation assumes small inter-frame displacement, the registration fails to converge under the motion in this sequence and tracking stops after 169 steps. RT-DLO produces per design no polyline; however, mask extraction takes 14.32 ms and sums up to 17.32 ms. To conclude,  MotionDLO reaches an complete speed of 12.34 ms, while both the mask extraction with 2.01 faster than RT-DLO with 14.32 and faster in the CPD step with 3.16 ms than TrackDLO.  For the mask extraction it is important to note that MotionDLO focuses only on moving wires and therefore doesn't take overlapping into account such as RT-DLO. The event-based CPD with 6.71 ms is faster than TrackDLO with 15.22 ms. For the real-time deployment of MotionDLO in a control loop it is important to adjust the time step updates accordingly to the inference time. 
\begin{table*}[htbp]
\caption{Inference time in ms of the MotionDLO pipeline decomposed by processing stage.}
\begin{center}
\begin{tabular}{|c|c|c|c|c|c|c|}
\hline
\textbf{Inference time in ms} & \textbf{\textit{Recentness map}} & \textbf{\textit{ROI masking}} & \textbf{\textit{Mask extraction}} & \textbf{\textit{Skeletonization}} & \textbf{\textit{CPD}} & \textbf{\textit{Complete}} \\
\hline
MotionDLO & 0.32 & 2.07 & 2.01 & 3.68 & 3.16 & 12.34  \\
\hline
 RT-DLO& -& 3.20& 14.32& -& -&-\\\hline
 TrackDLO& -& -& -& (669)& 15.22&-\\\hline
\end{tabular}
\label{tab:inferencetime}
\end{center}
\end{table*}

\subsubsection{Tracking Accuracy across DLO Types and Robot Speeds}
 
This tracking accuracy of MotionDLO is evaluated across different DLO types and robot speeds, and its performance is compared against TrackDLO on Dataset~2. Accuracy is reported using two complementary metrics: the average arc-length paired per-node error $e_t$ from~\eqref{eq:per_node_error}, and the point-to-curve error $e_t^{\mathrm{ptc}}$ from~\eqref{eq:ptc_error}.

Ground-truth centerlines are generated from different image sources depending on the evaluated tracker and motion regime. For the TrackDLO comparison, RealSense D415 RGB frames are extracted from the recorded bag files so that the tracker output and ground-truth polyline are represented in the same image coordinate system. An additional robot speed~$5\%$ recording is used for this comparison, since TrackDLO diverges at higher speeds~\cite{xiang_trackdlo_2023}. Physical tape markers were additionally used in this recording for visualization purposes only and were not used as a reference signal.

For MotionDLO, ground truth is generated from two sources. For the frame-based branch, IDS~uEye+ frames from the synchronized recordings are used. Although the camera is capable of $49\,\mathrm{fps}$, it was operated at $20\,\mathrm{fps}$ due to hardware trigger coupling with the event camera. This does not limit MotionDLO, because the frame-based branch is used only during static DLO states for initialization, while active manipulation is handled by the event-based branch. For the event-based branch, ground truth is generated from event-reconstruction frames obtained by accumulating ON/OFF events, with tracker updates produced at approximately $10\,\mathrm{ms}$ intervals.

For both TrackDLO and MotionDLO evaluations, the relevant frames are annotated in CVAT~\cite{cvat} using SAM~2~\cite{ravi_sam2_2024} with positive and negative point prompts, followed by manual correction where needed. Each binary mask is converted into an $M$-node ground-truth polyline using the skeletonization and B-spline resampling procedure described in Section~\ref{polyline fitting}, saved as an $M \times 2$ node array together with its timestamp. Tracker outputs follow the same file structure, enabling unified timestamp-based matching between tracker predictions and ground-truth centerlines.

TrackDLO is evaluated at robot speeds $5\%$ and $50\%$, at fixed $30\,\mathrm{ms}$ query intervals within the temporal overlap between the ground-truth and tracker streams, with the nearest ground-truth frame and tracker update paired by timestamp at each query step where its small-displacement assumption remains testable before failure~\cite{xiang_trackdlo_2023}.

\begin{figure}
    \centering
    \includegraphics[width=1\linewidth]{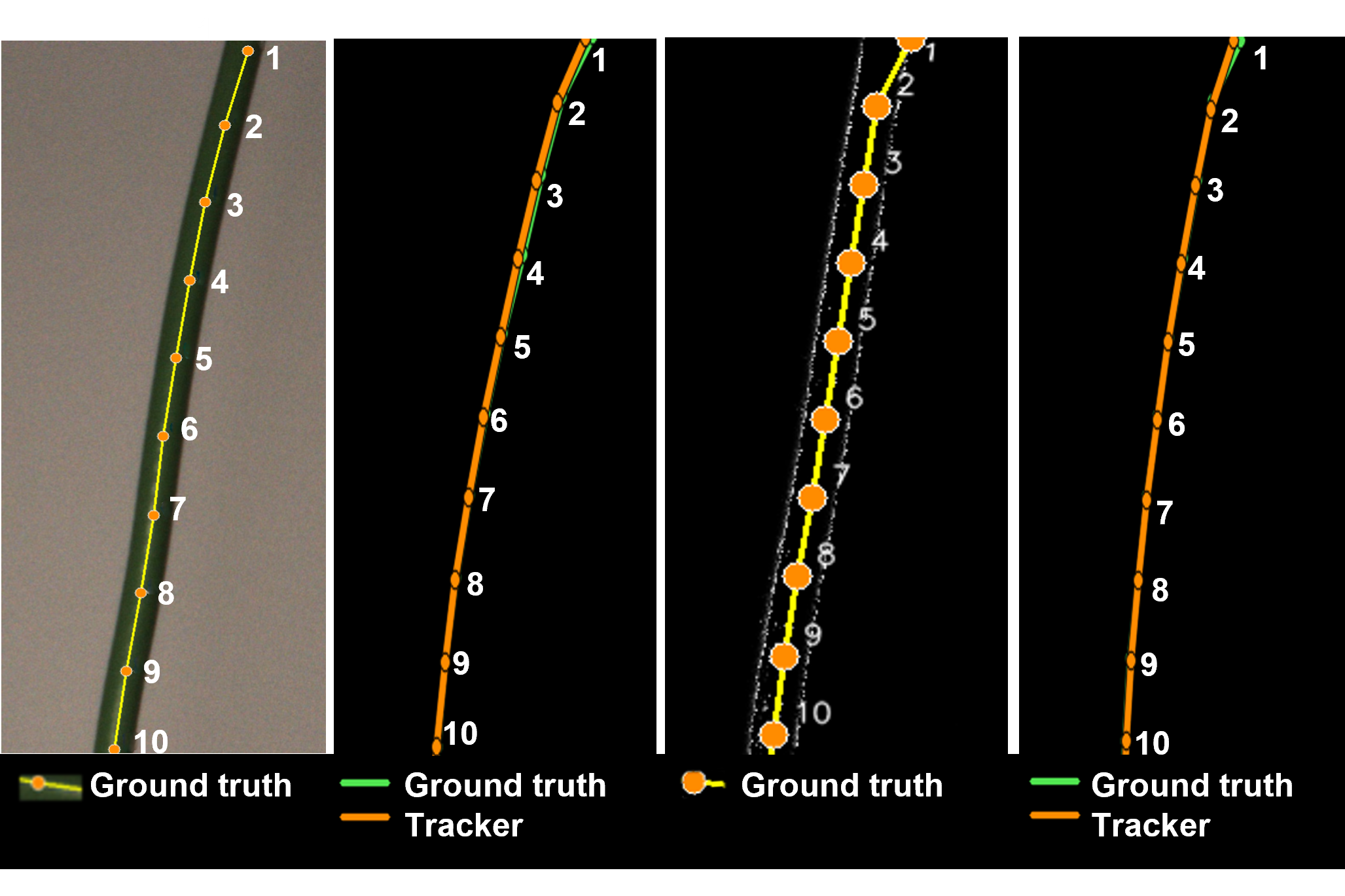}
    \caption{Ground truth generated for DLO 2 and its tracking result. Left: Frame based ground truth generation based on images at the beginning of the movement without motion blur and its respective tracking result. Right: Event-based ground truth generation based on the spare representation and its tracking result. For visualization purposes only 10 of the used 20 nodes are displayed.}
    \label{fig:GT_Tracker}
\end{figure}

MotionDLO is evaluated primarily at robot speed~$100\%$, which corresponds to the fast-motion regime targeted by the proposed method. Additional evaluations at robot speeds $5\%$ and $50\%$ are performed to enable direct comparison with TrackDLO. For MotionDLO, the evaluation grid is determined by the frame-based or event-based tracker update steps rather than by a fixed clock. Each tracker polyline is matched to the nearest available ground-truth polyline by timestamp before computing the error.

All errors are converted from pixels to millimetres using a per-DLO 
pixel-to-metric calibration. The scale is derived directly from the 
imaged DLO by matching the median cable width $w_{\mathrm{px}}$ measured 
from the segmentation mask to the caliper-measured outer diameter $d$:
\begin{equation}
  s = \frac{w_{\mathrm{px}}}{d}.
  \label{eq:pxmm_scale}
\end{equation}
Pixel-level errors are converted to metric units using 
$e_{\mathrm{mm}} = e_{\mathrm{px}}/s$, ensuring that all reported errors 
correspond to the physical work-surface plane and are independent of the 
RGB-to-event homography $\mathbf{H}_{\mathrm{rgb}\rightarrow\mathrm{ev}}$. 
For DLO~1, $w_{\mathrm{px}} = 33.0$\,px and $d_1 = 6$\,mm give 
$s_1 = 5.50$\,px/mm. For DLO~2, $w_{\mathrm{px}} = 34.0$\,px and 
$d_2 = 6.5$\,mm give $s_2 = 5.23$\,px/mm. For DLO~3, $w_{\mathrm{px}} = 72.5$\,px and 
$d_3 = 4$\,mm give $s_3 = 18.13$\,px/mm.

\begin{figure}
    \centering
    \includegraphics[width=\columnwidth]{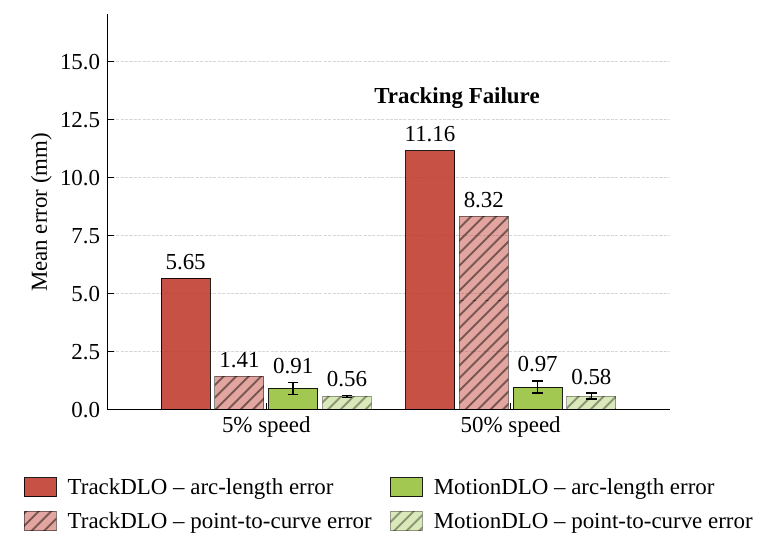}
    \caption{TrackDLO and MotionDLO compared at robot speeds 5 $\%$ and 50 $\%$. At robot speed 50 $\%$ TrackDLO starts to fail.}
    \label{fig:Comparison}
\end{figure}

\begin{figure}
    \centering
    \includegraphics[width=\columnwidth]{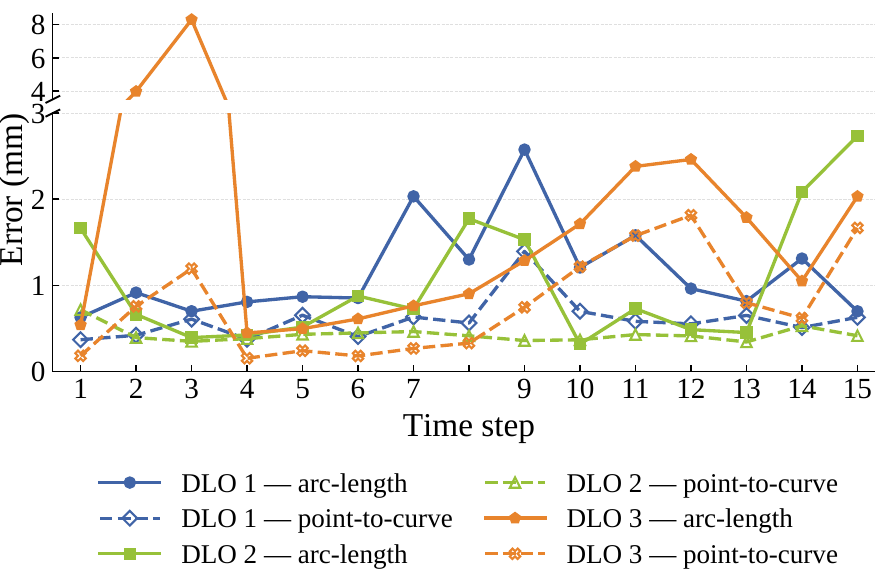}
    \caption{Tracking error at defined timesteps of MotionDLO for all three DLOs under the moving condition at robot speed 100 $\%$.}
    \label{fig:Point to Curve Error}
\end{figure}

Figure~\ref{fig:Comparison} summarizes TrackDLO's tracking accuracy on DLO 2. 
At the lower manipulation robot speed 5 $\%$, TrackDLO tracks the 
stationary-to-slow-moving DLO reliably, achieving $\bar{e} = 5.65$\,mm and 
$\bar{e}^{\mathrm{ptc}} = 1.41$\,mm, indicating stable and consistent 
tracking. At speed 50, however, tracking degrades 
immediately upon motion onset, with $\bar{e}$ rising to 11.16\,mm and $\bar{e}^{\mathrm{ptc}}$ increasing to 8.32\,mm, before failing to track. This confirms the small inter-frame displacement limitation of TrackDLO identified in~\cite{xiang_trackdlo_2023} and motivates the event-based branch of MotionDLO, which adapts to large DLO deformations without prior parameter tuning or task-specific training data.

At robot speed~5 $\%$, MotionDLO achieves $\bar{e} = 0.91$\,mm (std $0.25$\,mm) and $\bar{e}^{\mathrm{ptc}} = 0.56$\,mm (std $0.05$\,mm) over 15 steps, with a tracker-side temporal residual of $2.00$\,ms. At robot speed~50 $\%$, MotionDLO achieves $\bar{e} = 0.97$\,mm (std $0.26$\,mm) and $\bar{e}^{\mathrm{ptc}} = 0.58$\,mm (std $0.12$\,mm) over 16 evaluation steps, with a temporal residual of $0.00$\,ms. MotionDLO therefore reduces the arc-length paired error by approximately 6 times at robot speed~5 $\%$ and by more than an order of magnitude at robot speed~50 $\%$, while also tracking through the motion phase at which TrackDLO diverges. MotionDLO's accuracy is essentially unchanged between the two speeds, indicating that it is insensitive to the small-displacement assumption that bounds TrackDLO

\begin{figure}
    \centering
    \includegraphics[width=\columnwidth, trim={0 0 0 3mm, clip}]{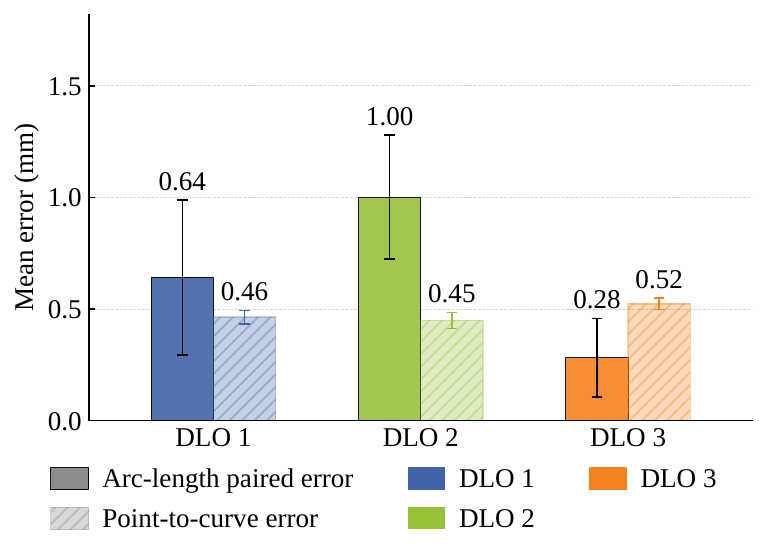}
    \caption{The mean arc-length paired and point-to-curve error of MotionDLO under a static condition for all three DLOs.}
    \label{fig:staticcondition}
\end{figure}

The geometric tracking accuracy of MotionDLO in the fast-motion regime targeted by the proposed method was further on $15$ evaluation steps per DLO drawn from the stable-motion phase at robot speed~100 $\%$. The tracker-side temporal residual measured against the asynchronous ground-truth timestamps is $0.00$\,ms for all three DLOs, in both mean and maximum, confirming that the hybrid event-frame pipeline preserves strict temporal alignment between the predicted centerline states $\hat{\mathbf{Y}}(t_k)$ and the ground-truth queries $\mathbf{Y}^{\mathrm{gt}}(t_k)$.

Quantitative results are summarised in Table~\ref{tab:tracking_accuracy}. Under the arc-length paired error ~\eqref{eq:per_node_error} of MotionDLO achieves $\bar{e} = 1.06$\,mm on DLO~1 , $\bar{e} = 1.02$\,mm on DLO~2 and $\bar{e} = 1.92$\,mm on DLO 3. The complementary point-to-curve metric of~\eqref{eq:ptc_error}, which is robust to local re-parameterisation of the predicted centerline and isolates geometric drift from arc-length mis-registration, yields substantially tighter values of $\bar{e}^{\mathrm{ptc}} = 0.55$\,mm on DLO~1 , $0.43$\,mm on DLO~2 and $\bar{e}^{\mathrm{ptc}} = 0.78$\,mm on DLO 3

The reduction from the arc-length to the point-to-curve metric is $48\%$ on DLO~1,$58\%$ on DLO~2 and $59\%$ on DLO~3, indicating that the dominant residual component is a tangential misalignment of node correspondences along the cable rather than a normal-direction deviation of the tracked curve from the true DLO geometry, a behaviour consistent with the CPD M-step bias toward smooth, low-curvature displacement fields under the MCT geodesic kernel.

Both metrics remain well below the respective cable diameters $d_1 = 6\,\mathrm{mm}$ (DLO~1), $d_2 = 6.5\,\mathrm{mm}$ (DLO~2) and  $d_3 = 4\,\mathrm{mm}$ (DLO~3) , with the point-to-curve mean corresponding to $9.2\%$ of the diameter on DLO~1 and $6.6\%$ on DLO~2, and $19.5\%$ on DLO~3, demonstrating that MotionDLO maintains sub-diameter geometric fidelity throughout the evaluated motion segment; the comparatively larger relative value on DLO~3 is driven primarily by two transient frames in which the tracked estimate diverged toward spurious mask fragments disconnected from the cable body.

Per-step errors across the evaluation window are shown in figure~\ref{fig:Point to Curve Error}.

\begin{table}[t]
  \centering
  \caption{MotionDLO tracking accuracy on DLO~1, DLO~2, and DLO~3
           ($n=15$ frames, robot speed~100 $\%$).
           Scales: $s=5.50\,\mathrm{px/mm}$ (DLO~1),
           $s=5.23\,\mathrm{px/mm}$ (DLO~2),
           $s=\text{18.12}\,\mathrm{px/mm}$ (DLO~3).}
  \label{tab:tracking_accuracy}
  \renewcommand{\arraystretch}{1.2}
  \begin{tabular}{llcccc}
    \toprule
    DLO & Metric
      & Mean $\bar{e}$
      & Median
      & Std $\sigma$
      & Unit \\
    \midrule
    \multirow{4}{*}{DLO~1}
      & \multirow{2}{*}{Arc-length paired}
        & $5.83$ & $4.64$ & $2.69$ & px \\
      & & $1.06$& $0.84$& $0.49$& mm \\
    \cmidrule{2-6}
      & \multirow{2}{*}{Point-to-curve}
        & $3.05$ & $2.94$ & $1.20$ & px \\
      & & $0.55$& $0.54$& $0.22$& mm \\
    \midrule
    \multirow{4}{*}{DLO~2}
      & \multirow{2}{*}{Arc-length paired}
        & $5.35$ & $3.78$ & $3.76$ & px \\
      & & $1.02$ & $0.72$ & $0.72$ & mm \\
    \cmidrule{2-6}
      & \multirow{2}{*}{Point-to-curve}
        & $2.24$ & $2.15$ & $0.47$ & px \\
      & & $0.43$ & $0.41$ & $0.09$ & mm \\
    \midrule
    \multirow{4}{*}{DLO~3}
      & \multirow{2}{*}{Arc-length paired}
        & 34.72&23.30& 35.34 & px \\
     & & 1.92&1.29& 1.95& mm \\
    \cmidrule{2-6}
      & \multirow{2}{*}{Point-to-curve}
        & 14.15& 13.47& 10.2 & px \\
    & & 0.78& 0.74& 0.56& mm \\
    \bottomrule
  \end{tabular}
\end{table}

As the frame-based branch of MotionDLO provides the geometric reference used to initialise and periodically re-initialise the event-based tracker during static phases, it is evaluated separately, against frame-based ground truth images, on nine evaluation steps per DLO drawn from the stationary and pre-motion phase of three independent runs at Speed~100. The mean arc-length paired and point-to-curve errors for all three DLOs are summarised in figure .~\ref{fig:staticcondition}. For DLO~1, the frame-based branch achieves $\bar{e} = 0.64$\,mm and $\bar{e}^{\mathrm{ptc}} = 0.46$\,mm. For DLO~2, the corresponding values are $\bar{e} = 1.00$\,mm and $\bar{e}^{\mathrm{ptc}} = 0.45$\,mm, closely matching the event-based moving results of $1.02$\,mm and $0.43$\,mm on the same DLO and confirming that the two branches produce geometrically consistent centerline estimates. For DLO~3, $\bar{e} = 0.28$\,mm and 
$\bar{e}^{\mathrm{ptc}} = 0.52$\,mm, representing the lowest arc-length error across all three cables and confirming accurate shape recovery on the thinnest 4.0\,mm cable. Across all three DLOs, 
both metrics remain well within the sub-diameter regime, confirming that the frame-based branch provides a reliable initial state for the 
event-based tracker.

\subsection{Limitations and Failure Modes}

The current formulation assumes that a single DLO is actively manipulated at any instant. Dataset~1, for example, contains four mutually overlapping DLOs, of which only one is driven. For initialization a static condition is assumed. When two or more DLOs move simultaneously or when the manipulated DLO drags a quiescent neighbor into motion through frictional contact, the event-derived
motion mask cannot disambiguate the co-moving instances, since both contribute
events to the same spatial neighborhood. SAM~3 consequently segments the
overlapping strands as a single instance, and the single-instance CPD tracker
locks its node set onto this merged mask. Once the
node correspondences span two physical DLOs, the registered centerline no
longer corresponds to any single object. Resolving this case requires either
instance-level event clustering prior to segmentation or a multi-instance
tracking back end, both of which we leave to future work. The recentness map presupposes that the static background generates no events, so that all activity in the event stream can be attributed to DLO motion. This assumption is violated whenever the scene contains independent event sources within the field of view, such as moving conveyor belts, vibrating fixtures, or other robots. The spurious events they emit can be similar, at the level of the recentness map, from cable-induced events, and they therefore contaminate the motion mask and bias the segmentation prompt. In structured cells this can be mitigated by spatial masking of known dynamic regions, but no such suppression is currently part of the pipeline. MotionDLO recovers a 2D projection of the DLO centerline onto the calibrated work plane; it does not estimate out-of-plane shape. Configurations in which the DLO lifts substantially off the plane, or self-occludes in depth, are therefore not represented. Full 3D shape recovery would require depth sensing or multi-view fusion, a natural extension given the existing stereo calibration between the frame and event cameras. 

\section{Conclusion and Future Work}

\subsection{Conclusion}
This paper presents MotionDLO, a hybrid event- and frame-based tracking framework for DLOs that addresses the fundamental trade-off between temporal resolution and spatial accuracy in existing DLO perception pipelines. By combining the high temporal responsiveness of event-based sensing with the geometric generalizability of zero-shot SAM 3 segmentation, the proposed method enables robust shape estimation under fast motion conditions where purely frame-based approaches fail.
 
The core of the framework is a modified CPD tracker that operates on event-derived observations at 12 ms update intervals, augmented with five targeted extensions: chain Laplacian regularization to enforce topology-aware smoothness along the one-dimensional DLO structure, an inter-frame velocity warm-start that propagates motion estimates through the geodesic kernel, noise variance management strategy that prevents both numerical instability and unbounded growth of~$\sigma^2$, a two-pass registration scheme with Gaussian soft observation weighting that replaces hard spatial pruning, and an adaptive re-initialization mechanism that recovers from tracking loss without manual intervention. 
 
Experimental evaluation on a dataset of 54 measurement sequences with 1,858 frames and 6.2 GB of events spanning three industrially representative DLO types and three manipulation speed levels demonstrated that MotionDLO achieves an average point-to-curve error of 0.54 mm while maintaining an inference time of 12 ms per event update-well within the requirements for closed-loop robotic control. In comparative benchmarks, MotionDLO eliminates identity switches that are consistently observed in RT-DLO and reduces tracking failures under fast motion relative to TrackDLO, which is unable to operate on DLOs with diameters below 6 mm due to depth-sensing limitations. The temporally coherent correspondence estimation inherent to the CPD formulation prevents the re-identification failures that affect per-frame methods in dense multi-DLO environments, enabling stable long-horizon tracking across all tested scenarios.
 
In contrast to SOTA algorithms, the proposed approach operates without task-specific training data, relying instead on the zero-shot generalization capability of SAM 3 and the model-free registration properties of CPD. This design choice ensures applicability across diverse DLO types without retraining or domain adaptation, a practical advantage for deployment in industrial settings where cable types vary frequently. Although the experimental evaluation is conducted on DLOs ranging from 4 mm single-core cable to 8 mm pneumatic tubing, the method is not restricted to this diameter range; applicability to thinner or thicker objects is achieved by scaling the camera's FOV accordingly.

\subsection{Future Work} \label{Future Work}

Several directions for future investigation arise. First, the frame-based segmentation branch currently relies on SAM 3 inference, which constrains the correction rate to approximately 1 s. Replacing or complementing this component with a more lightweight segmentation model could increase the correction frequency while reducing computational overhead, further tightening the drift bound of the event-based tracker. Second, the current fusion mechanism employs confidence- and timestamp-weighted blending. A more principled probabilistic formulation, for instance, treating the two branches as heterogeneous observations within a unified Bayesian state estimator, could yield improved uncertainty quantification and more graceful degradation under partial sensor failure. Third, extending the framework to three-dimensional DLO tracking by incorporating depth information from stereo event cameras or structured-light sensors would broaden applicability to manipulation tasks that require out-of-plane shape estimation. Fourth, the present evaluation assumes approximately constant background illumination; investigating robustness under rapidly varying lighting conditions, such as those encountered in welding or laser-cutting environments, remains an open challenge. Finally, integrating the tracker into a closed-loop manipulation pipeline with online path adaptation would provide a direct assessment of its utility for robotic DLO handling, the motivating application of this work.

\section*{Acknowledgment}

The results presented in this paper were developed as part of the project PES4E|Road - Power electronic systems for electrified roads, funded by the European Regional Development Fund (EFRE) through ‘Investment in Employment and Growth’ Bavaria 2021-2027. Any opinions, findings, and conclusions or recommendations expressed in this paper are those of the author and do not necessarily reflect the views of the partners. The authors acknowledge the use of Anthropic Claude to assist with language editing and improving the readability of the manuscript text. All AI-generated suggestions were reviewed, verified, and revised by the authors, who bear full responsibility for the content.

\printbibliography
\end{document}